\documentclass{LMCS}

\usepackage{amsmath}
\usepackage{amsfonts}
\usepackage{amssymb}
\usepackage[noaddress]{chicagor}
\usepackage{hyperref}

\allowdisplaybreaks[2]
\newcommand{\pro}{\begin{prop}}
\newcommand{\epro}{\end{prop}}
\newcommand{\prf}{\proof}
\newcommand{\eprf}{\qed}

\newenvironment{oldtheorem}[1]
  {\begin{renewcommand}{\thethm}{\ref{#1}}}
  {\end{renewcommand}\addtocounter{thm}{-1}}

\newcommand{\Rule}[2]{          %
  \begin{array}{c}
  #1 \\\hline
  #2
  \end{array}}

\newcommand{\sat}{\models}
\newcommand{\rimp}{\Rightarrow}
\renewcommand{\phi}{\varphi}

\renewcommand{\emptyset}{\varnothing}
\newcommand{\riff}{\Leftrightarrow}

\newcommand{\ob}{\mathit{ob}}

\newcommand{\union}{\ensuremath{\cup}}

\newcommand{\cH}{\mathcal{H}}
\newcommand{\cO}{\mathcal{O}}
\newcommand{\cL}{\mathcal{L}}
\newcommand{\cP}{\mathcal{P}}
\newcommand{\cF}{\mathcal{F}}
\newcommand{\cK}{\mathcal{K}}
\newcommand{\cM}{\mathcal{M}}
\newcommand{\cE}{\mathcal{E}}
\newcommand{\cW}{\mathcal{W}}

\newcommand{\N}{N}

\renewcommand{\Pr}{\mathrm{Pr}}
\newcommand{\Ev}{\mathrm{Ev}}
\newcommand{\lEv}{\underline{\mathrm{Ev}}}
\newcommand{\uEv}{\overline{\mathrm{Ev}}}

\newcommand{\true}{\mbox{\textbf{true}}}
\newcommand{\false}{\mbox{\textbf{false}}}

\newcommand{\COMMENTOUT}[1]{}

\renewcommand{\ng}{r}

\newcommand{\A}{\mathtt{A}}
\newcommand{\Ad}{\mathtt{A}^d}
\newcommand{\YES}{\mbox{``Yes''}}
\newcommand{\NO}{\mbox{``No''}}
\newcommand{\DONTK}{\mbox{``?''}}
\newcommand{\pprime}{\mathsf{prime}}
\newcommand{\pwall}{\mathsf{wall}}
\newcommand{\pdhead}{\mathsf{dh}}

\newcommand{\AAlice}{\A_{\mathrm{Alice}}}
\newcommand{\ARobot}{\A_{\mathrm{Robot}}}
\newcommand{\ABob}{\A_{\mathrm{Bob}}}

\newcommand{\wlow}{\underline{w}}
\newcommand{\wup}{\overline{w}}

\newcommand{\commentout}[1]{}

\def\labelenumi{{\rm(\arabic{enumi})}}

\def\doi{1 (3:1) 2005}
\lmcsheading%
{\doi}
{26}
{}
{}
{Mar.~20, 2005}
{Dec.~20, 2005}
{}

\begin{document}

\title[Probabilistic Algorithmic Knowledge]{Probabilistic Algorithmic Knowledge}

\author[J.~Y.~Halpern]{Joseph Y.~Halpern\rsuper a} 
\address{{\lsuper a}Cornell University, Ithaca, NY 14853 USA} 
\email{halpern@cs.cornell.edu}  

\author[R.~Pucella]{Riccardo Pucella\rsuper b}   
\address{{\lsuper b}Northeastern University, Boston, MA 02115 USA} 
\email{riccardo@ccs.neu.edu}

\keywords{Knowledge, Probability, Evidence, Randomized Algorithms,
  Algorithmic Knowledge, Logic}
\subjclass{I.2.4; G.3}

\maketitle

\begin{abstract}
The framework of algorithmic knowledge assumes that agents use
deterministic knowledge algorithms to compute the facts they
explicitly know. 
We extend the framework to allow for randomized
knowledge algorithms. 
We then
characterize the information provided by a randomized knowledge algorithm
when its answers have some probability of being incorrect.  We formalize
this information in terms of {\em evidence};
a randomized knowledge algorithm returning
``Yes'' to a query about a fact $\phi$ provides evidence for $\phi$
being true. 
Finally, we discuss the extent to which this evidence can be used as a
basis for decisions.  
\end{abstract}

\section{Introduction}

Under the standard possible-worlds interpretation of knowledge, which
goes back to Hintikka \citeyear{Hi1}, 
an agent knows $\phi$ if $\phi$ is true at all the worlds the agent
considers possible.
This interpretation of knowledge has been found useful in capturing some
important intuitions in the analysis of distributed protocols
\cite{r:fagin95}.  
However, its usefulness is somewhat limited by what
Hintikka \citeyear{Hi1} called the {\em logical omniscience
problem\/}:~agents know all tautologies and know all logical
consequences of their knowledge.   
Many approaches have been developed to deal with the logical omniscience
problem (see 
\cite[Chapter 10 and 11]{r:fagin95} for a discussion and survey).  We
focus on one approach here that has been called 
\emph{algorithmic knowledge} \cite{r:halpern94}.
The idea is simply to assume that agents 
are equipped with ``knowledge algorithms'' that they use to
compute what they know.  
An agent {\em algorithmically knows\/} $\phi$ if his knowledge algorithm
says ``Yes'' when asked $\phi$.%
\footnote{We remark that what we are calling ``knowledge algorithms''
here are called ``local algorithms'' in \cite{r:fagin95,r:halpern94}.}

Algorithmic knowledge is a very general approach.
For example, Berman, Garay, and Perry \citeyear{r:berman89} implicitly use a
particular form of algorithmic knowledge in their analysis of Byzantine
agreement.  Roughly speaking they allow agents to perform limited tests
based on the information they have; agents know only what follows from
these limited tests.  
Ramanujam \citeyear{r:ramanujam99} investigates a particular form of 
algorithmic knowledge, where the knowledge algorithm is essentially a
model-checking procedure for a standard logic of knowledge. More
specifically, Ramanujam considers, at every state, the part of the
model that a particular agent sees (for instance, an agent in a
distributed system may be aware only of its immediate neighbors, the ones with
whom he can communicate) and takes as knowledge algorithm the
model-checking procedure for epistemic logic, applied to the submodel
generated by the visible states. 
Halpern and Pucella \citeyear{r:halpern02e} have applied algorithmic
knowledge to security
to capture adversaries who are resource bounded (and thus, for example, 
cannot factor the products of large primes that arise in the
RSA cryptosystem \cite{RSA}).
\COMMENTOUT{
Finally, an important area of applicability for algorithmic knowledge
is in knowledge-based programming. Roughly speaking, knowledge-based
programs \cite{FHMV94} are programs that has tests for knowledge. For
instance, one can write a program that essentially says ``if you know
that there is obstacle in front of you, turn right, otherwise,
continue.'' This kind of program is very high-level (one does not need
to know exactly how knowledge of the fact that there is an obstacle
comes to be), and captures nicely the intuition underlying some
classes of programs. For examples in robotics, see \cite{BLMS}. A
feature of knowledge-based programs is that they cannot be directly
executed. A typical approach is to ``implement'' the knowledge-based
program using a so-called standard program, one that does not make
reference to knowledge. It is not always possible, to implement a
knowledge-based program via a standard program, and in some cases,
there is not a unique standard program that implements a knowledge
program. 
An alternative to implementing knowledge-based programs via a standard
program, however, is to replace every test for knowledge in a
knowledge-based program by a test for algorithmic knowledge, that is,
a call to the knowledge algorithm of the agent, to evaluate knowledge
queries.
}
All these examples use {\em sound\/} knowledge algorithms:~although the
algorithm may not give an answer under all circumstances, when it says
``Yes'' on input $\phi$, the agent really does know $\phi$ in the
standard possible-worlds sense.  Although soundness is not required in
the basic definition, it does seem to be useful in many applications.

Our interest in this paper is knowledge algorithms that may use some
randomization.  As we shall see, there are numerous examples of natural
randomized knowledge algorithms.
With randomization, whether or not the knowledge
algorithm says ``Yes'' may depend on the outcome of coin tosses.
This poses a slight difficulty in even giving semantics to algorithmic
knowledge, since the standard semantics makes sense only for
deterministic algorithms.  To deal with this problem,
we make the
algorithms deterministic by supplying them an extra argument
(intuitively, the outcome of a sequence of coin tosses) to
``derandomize'' them.  We show that this approach provides a natural
extension of the deterministic case.

To motivate the use of randomized knowledge algorithms, we consider a
security example from Halpern and Pucella \citeyear{r:halpern02e}. 
The framework in that paper lets us reason 
about principals communicating in the presence of adversaries, using
cryptographic protocols.  
The typical 
assumption made when analyzing security protocols is that adversaries
can intercept all the messages exchanged by the principals, but cannot
necessarily decrypt encrypted messages unless they have the
appropriate decryption key. To capture precisely the capabilities of
adversaries, we use knowledge algorithms. Roughly speaking, a
knowledge algorithm for an adversary will specify what information the
adversary can extract from intercepted messages. In this paper, we
consider an adversary that further attempts to guess the cryptographic
keys used by the principals in the protocol. We show how to capture
the knowledge of such an adversary using a randomized knowledge
algorithm.

Having defined the framework, we try to characterize the
information obtained by getting a $\YES$ answer to
a query for $\phi$. If the knowledge algorithm is sound, 
then a $\YES$ answer guarantees that $\phi$ is true.
However, the randomized algorithms of most interest to us 
give wrong answers with positive probability, so are 
not sound.  Nevertheless, it certainly seems
that if the probability that the algorithm gives the wrong answer is
low, it 
provides very useful information when it says ``Yes'' to a query
$\phi$. 
This intuition already appears in the randomized algorithms
literature, where a   
$\YES$ answer from a highly reliable randomized algorithm
(i.e., one  with a low probability of being wrong) is
deemed ``good enough''. 
In what sense is this true?  One contribution of our
work is to provide a formal answer to that question. 
It may seem that a $\YES$ answer to a query $\phi$ from a highly
reliable randomized knowledge algorithm should make the probability 
that $\phi$ is true be high but, as
we show, this is not necessarily true.  Rather, the information should
be viewed as {\em evidence\/} that $\phi$ is true; 
the probability that $\phi$ is true also depends in part on the prior
probability of $\phi$.

Evidence has been widely studied in the literature on inductive logic
\cite{r:kyburg83}. 
We focus on the evidence
contributed specifically by a randomized knowledge algorithm. 
In a companion paper \cite{r:halpern03b}, we consider a formal logic for
reasoning about evidence.
The rest of this paper is organized as follows.  In
Section~\ref{s:algknow}, we review algorithmic knowledge (under the
assumption that knowledge algorithms are deterministic).  In
Section~\ref{s:random}, we give semantics to algorithmic knowledge in
the presence of randomized knowledge algorithms.  In
Section~\ref{s:security}, we show how the definition works in the
context of an example from the security domain.  In
Section~\ref{s:probalgknow} we characterize the information provided by
a randomized knowledge algorithm in terms of evidence.  We conclude in
Section~\ref{s:conc}.  All proofs are deferred to the appendix.

\section{Reasoning about Knowledge and Algorithmic Knowledge}\label{s:algknow}

The aim is to be able to reason about properties of systems involving
the knowledge of agents in the system. To formalize this type of
reasoning, we first need a language. The syntax for a multiagent logic
of knowledge is straightforward. Starting with a set $\Phi$ of
primitive propositions, which we can think of as describing basic
facts about the system, such 
as
``the door is closed'' or ``agent $A$
sent the message $m$ to $B$'', more complicated formulas are formed by
closing off under negation, conjunction, and the modal operators
$K_1$, $\ldots$, $K_n$ and $X_1,\ldots,X_n$. Thus, if $\phi$ and
$\psi$ are formulas, then so are $\neg\phi$, $\phi\wedge\psi$,
$K_i\phi$ (read ``agent $i$ knows $\phi$''), and $X_i\phi$ (read
``agent $i$ can compute $\phi$''). 
As usual, we take $\phi\vee\psi$ to be an abbreviation for
$\neg(\neg\phi\wedge\neg\psi)$ and $\phi\rimp\psi$ to be an
abbreviation for $\neg\phi\vee\psi$.

The standard possible-worlds semantics for knowledge uses {\em Kripke
structures} \cite{r:kripke63}.  Formally, a
Kripke structure is composed of a set $S$ of states or possible
worlds, an interpretation $\pi$ which associates with each state in
$S$ a truth assignment to the primitive propositions (i.e., $\pi(s)(p)
\in \{\true, \false\}$ for each state $s\in S$ and each primitive
proposition $p$), and equivalence relations $\sim_i$ on $S$ (recall
that an equivalence relation is a binary relation which is reflexive,
symmetric, and transitive). The relation $\sim_i$ is agent $i$'s
possibility relation. Intuitively, $s\sim_i t$ if agent $i$ cannot
distinguish state $s$ from state $t$ (so that if $s$ is the actual
state of the world, agent $i$ would consider $t$ a possible state of
the world). For our purposes, the equivalence relations are obtained
by taking a set $\cL$ of \emph{local states}, and giving each agent a
\emph{view} of the state, that is, a function $L_i : S
\rightarrow \cL$. We define $s\sim_i t$ if and only if $L_i(s)=L_i(t)$. In
other words, agent $i$ considers the states $s$ and $t$
indistinguishable if he has the same local state at both states.

To interpret explicit knowledge of the form $X_i\phi$, we assign to
each agent a \emph{knowledge algorithm} that the agent can use to
determine whether he knows a particular formula. A knowledge algorithm
$\A$ takes as inputs a formula of the logic, a local state $\ell$ in
$\cL$, as well as the state as a whole.
This is a generalization of the original presentation of algorithmic
knowledge \cite{r:halpern94}, 
in which the knowledge algorithms did not take the state as input. 
The added generality is necessary to model knowledge algorithms that
query the state---for example, a knowledge algorithm might use a
sensor to determine the distance between a robot and a wall (see
Section~\ref{s:probalgknow}). 
Knowledge algorithms are required 
to be deterministic and terminate on all inputs,
with result $\YES$, $\NO$, or $\DONTK$. 
A knowledge algorithm says $\YES$ to a formula $\phi$ (in a given
state) if the algorithm determines that the agent knows $\phi$ at the
state, $\NO$ if the algorithm determines that the agent does not know
$\phi$ at the state, and $\DONTK$ if the algorithm cannot determine
whether the agent knows $\phi$. 

An \emph{algorithmic 
knowledge 
structure} $M$ is a tuple
$(S,\pi,L_1,\ldots,L_n,\A_1,\ldots,\A_n)$, where $L_1,\ldots,L_n$ are
the view functions on the states, and $\A_1,\ldots,\A_n$ are
knowledge algorithms.\footnote{Halpern, Moses, and Vardi \citeyear{r:halpern94}
introduced algorithmic knowledge in the context of dynamic systems,
that is, systems evolving in time.  
The knowledge algorithm is allowed to change at every state of the
system. Since the issues that interest us do not involve time, we do
not consider dynamic systems 
in this paper.
We remark that what we are calling ``algorithmic knowledge structures''
here are called ``algorithmic structures'' in \cite{r:fagin95,r:halpern94}.  The
term ``algorithmic knowledge structures'' 
is used in the paperback edition of \cite{r:fagin95}.}

We define what it means for a formula $\phi$ to be true (or satisfied)
at a state $s$ in an algorithmic 
knowledge
structure $M$, written
$(M,s)\sat\phi$, inductively as follows:
\begin{itemize}
\item[] $(M,s)\sat p$ if $\pi(s)(p)=\true$
\item[] $(M,s)\sat\neg\phi$ if $(M,s)\not\sat\phi$
\item[] $(M,s)\sat\phi\wedge\psi$ if $(M,s)\sat\phi$ and $(M,s)\sat\psi$
\item[] $(M,s)\sat K_i\phi$ if $(M,t)\sat\phi$ for all $t$ with $s\sim_i t$
\item[] $(M,s)\sat X_i\phi$ if $\A_i(\phi,L_i(s),s)=\YES$.
\end{itemize}
The first clause shows how we use the $\pi$ to define the semantics of
the primitive propositions. The next two clauses, which define the
semantics of $\neg$ and $\wedge$, are the standard clauses from
propositional logic. The fourth clause is designed to capture the
intuition that agent $i$ knows $\phi$ exactly if $\phi$ is true in all
the states that $i$ considers possible. The final clause interprets
$X_i\phi$ via agent $i$'s knowledge algorithm. Thus, agent $i$ has
algorithmic knowledge of $\phi$ at a given state if the agent's
algorithm outputs $\YES$ when presented with $\phi$, the
agent's local state, and the state. (Both the outputs $\NO$ and
$\DONTK$ result in 
lack of algorithmic knowledge.)
As usual, we say that a formula $\phi$ is {\em valid in structure $M$\/}
and write $M \sat \phi$ 
if $(M,s) \sat \phi$ for all states $s \in S$; $\phi$ is {\em valid\/}
if it is valid in all structures.

We can think of $K_i$ as representing \emph{implicit knowledge}, facts
that the agent implicitly knows, given its information. One can check
that implicit knowledge is closed under implication, that is,
$K_i\phi\land K_i(\phi\rimp\psi)\rimp K_i\psi$ is valid, and that an
agent implicitly knows all valid formulas, so that if $\phi$ is valid,
then $K_i\phi$ is valid. These properties say that agents are very
powerful reasoners. What is worse, while it is
possible to change some properties of knowledge by changing the
properties of the relation $\sim_i$, no matter how we change it, we
still get closure under implication and knowledge of valid formulas as
properties. They seem to be inescapable features of the
possible-worlds approach. This suggests that the possible-worlds
approach is appropriate only for ``ideal knowers'', ones that know all
valid formulas as well as all logical consequences of their knowledge,
and thus inappropriate for reasoning about agents that are
computationally limited. 
In contrast, $X_i$ represents \emph{explicit knowledge}, facts whose
truth the agent can compute explicitly. Since we put no a priori
restrictions on the knowledge algorithms, an agent can explicitly
know both $\phi$ and $\phi\rimp\psi$ without explicitly knowing
$\psi$, for example.

As defined, there is no necessary connection between $X_i\phi$ and
$K_i\phi$. An algorithm could very well claim that agent $i$ knows
$\phi$ (i.e., output $\YES$) whenever it chooses to, including at
states where $K_i\phi$ does not hold. Although algorithms that make
mistakes are common, we are often interested in knowledge algorithms
that are correct. We say that a knowledge algorithm is \emph{sound}
for agent $i$ in the structure $M$ if for all states $s$ of $M$ and
formulas $\phi$, $\A_i(\phi,L_i(s),s)=\YES$ implies $(M,s)\sat K_i\phi$, and
$\A_i(\phi,L_i(s),s)=\NO$ implies $(M,s)\sat\neg K_i\phi$. Thus, a knowledge
algorithm is sound if its definite answers are 
correct. 
If we restrict attention to sound algorithms, then
algorithmic knowledge can be viewed as an instance of awareness, as
defined by Fagin and Halpern \citeyear{FH}.  
There is a subtlety here, due to the asymmetry in the handling of the
answers returned by knowledge algorithms. The logic does not let us
distinguish between a knowledge algorithm returning $\NO$ and a
knowledge algorithm returning $\DONTK$; they both result in lack of
algorithmic knowledge.\footnote{There may be reasons to distinguish
$\NO$ from $\DONTK$, and it is certainly possible to extend the logic
to distinguish them.} In 
section~\ref{s:reliable}, where we define the notion of
a reliable knowledge algorithm, reliability will be characterized in
terms of algorithmic knowledge, and thus the definition will not
distinguish between a knowledge algorithm returning $\NO$ or
$\DONTK$. 
Thus, in that section, for simplicity, we consider algorithms that are
\emph{complete}, in the sense that they always return either $\YES$ or
$\NO$, and not $\DONTK$. More precisely, for a formula $\phi$, define
a knowledge algorithm $\A_i$ to be \emph{$\phi$-complete} for agent
$i$ in the structure $M$ if for all states $s$ of $M$,
$\A_i(\phi,L_i(s),s)\in\{\YES,\NO\}$. 

\section{Randomized Knowledge Algorithms}\label{s:random}

{\em Randomized\/} knowledge algorithms arise frequently in the
literature (although they have typically not been viewed as knowledge
algorithms).  
In order to deal with randomized algorithms in our
framework, we need to address a technical question. Randomized
algorithms are possibly nondeterministic; they may not yield the same
result on every invocation with the same arguments. Since $X_i\phi$
holds at a state $s$ if the knowledge algorithm answers $\YES$ at that
state, this means that, with the semantics of the previous section,
$X_i \phi$ would not be well defined. 
Whether it holds at a given state depends on the outcome of random
choices made by the algorithm.
However, we expect the semantics to unambiguously declare a formula
either true or false.
Before we describe our solution to the problem, we discuss another
potential solution, which is to define the satisfaction relation
probabilistically.  That 
is, rather than associating a truth value with each formula at each
state, we associate a probability $\Pr_s(\phi)$ with each formula $\phi$
at each state $s$.  The standard semantics can be viewed as a special
case of this semantics, where the probabilities are always either 0 or 1.
Under this approach, it seems reasonable to take $\Pr_s(p)$ 
to be either 0 or 1, depending on whether
primitive proposition $p$ is true at state $s$, 
and to take $\Pr_s(X_i\phi)$ to be the
probability that $i$'s knowledge algorithm returns ``Yes'' given
inputs $\phi$, $L_i(s)$, and $s$.  However, it is not then clear how to define 
$\Pr_s(\phi \land \psi)$.  Taking it to be $\Pr_s(\phi) \Pr_s(\psi)$
implicitly treats $\phi$ and $\psi$ as independent, which is clearly
inappropriate if $\psi$ is $\neg \phi$.%
\footnote{To get around this particular problem, some approaches that
  combine logic and probability 
give semantics to formulas by viewing them as random variables (e.g.,
\cite{r:kozen85}).} 
Even ignoring this problem, it
is not clear how to define $\Pr_s(X_i \phi \land X_i \psi)$, since again
there might be correlations between the output of the knowledge
algorithm on input $(\phi,L_i(s),s)$ and input ($\psi,L_i(s),s)$.%

We do not use probabilistic truth values in this paper.  Instead,
we deal with the problem by adding information to the semantic model to
resolve the uncertainty
about the truth value of formulas of the form $X_i \phi$.  
Observe that if
the knowledge algorithm $\A$ is randomized, then the answer that $\A$
gives on input $(\phi, \ell, s)$ will depend on the outcome of coin
tosses (or whatever other randomizing device is used by $\A$).  
We thus
turn the randomized
algorithm into a deterministic algorithm by supplying it with an
appropriate argument.  For example, we supply an algorithm that makes
random choices by tossing coins a sequence of outcomes of coin tosses.
We can now interpret a knowledge algorithm answering $\YES$ with
probability $\alpha$ at a state by considering the probability of those
sequences of coin tosses at the state that make the algorithm answer
$\YES$. 
Formally, we start with (possibly randomized) knowledge
algorithms $\A_1,\ldots,\A_n$.  For
simplicity, assume that the randomness in the knowledge algorithms
comes from tossing coins. 
A \emph{derandomizer} is a tuple
$v=(v_1,\ldots,v_n)$ such that for every agent $i$, $v_i$ is a
sequence of outcomes of coin tosses (heads and tails).  There is a
separate sequence of coin tosses for each agent rather than just a
single sequence of coin tosses, since we do not want to assume that
all agents use the same coin.  Let $V$ be the set of all such
derandomizers.  To every randomized algorithm $\A$ we associate a
derandomized algorithm $\Ad$ which takes as input not just the query
$\phi$, local state $\ell$, and state $s$, but also the sequence $v_i$
of $i$'s 
coin tosses, taken from a derandomizer $(v_1, \ldots, v_n)$.  A
\emph{probabilistic algorithmic 
knowledge
structure} 
is a tuple $\N=(S,\pi,L_1,\ldots,L_n,\Ad_1,\ldots,\Ad_n,\nu)$, where
$\nu$ is a 
probability distribution on $V$ and $\Ad_i$ is the derandomized
version of $\A_i$. 
(Note that in a probabilistic algorithmic 
knowledge
structure the knowledge algorithms are in fact deterministic.)  

The only assumption we make about the distribution $\nu$ is that it does
not assign zero probability to the nonempty sets of sequences of coin
tosses that determine  the result of the knowledge algorithm. More
precisely, we assume that for all agents $i$, formulas $\phi$, and 
states $s$,
$\{v\mid\A^d_i(\phi,L_i(s),s,v_i)=\YES\}\ne\emptyset$ if and only if  
$\nu(\{v\mid \A^d_i(\phi,L_i(s),s,v_i)=\YES\})>0$, and similarly for
$\NO$ and $\DONTK$ answers. Note that this property is satisfied, for
instance, if $\nu$ assigns nonzero probability to every 
sequence of coin tosses. 
We do not impose any other restrictions on $\nu$.  In particular, 
we do not require that the coin 
be fair or that the tosses be independent. Of course, we can 
capture correlation between the agents' coins by using an appropriate
distribution $\nu$. 

The truth of a formula is now determined relative to a pair $(s,v)$
consisting of a state $s$ and a derandomizer $v$.  We abuse notation
and continue to call these pairs states.  
The semantics of formulas in a probabilistic algorithmic 
knowledge 
structure is a straightforward extension of their semantics in
algorithmic 
knowledge
structures. The semantics of primitive propositions is given by $\pi$;
conjunctions and negations are interpreted as usual; for knowledge and
algorithmic knowledge, we have
\begin{itemize}
\item[]$(\N,s,v) \sat K_i\phi$ if $(\N,t,v')\sat\phi$ for all $v'\in
V$ and all $t\in S$ such that $s\sim_i t$%
\item[]$(\N,s,v) \sat X_i\phi$ if $\Ad_i(\phi,L_i(s),s,v_i)=\YES$, where $v=(v_1,\ldots,v_n)$.
\end{itemize}
Here, $\Ad_i$ gets $v_i$ as part of its input.  $\Ad_i(\phi,L_i(s),s,v_i)$ is
interpreted as the output of $\Ad_i$ given that $v_i$ describes the
outcomes of the coin tosses.  
It is perhaps best to interpret $(M,s,v) \sat X_i \phi$ as saying that
agent $i$'s knowledge algorithm would say ``Yes'' if it were run in
state $s$ with derandomizer $v_i$.  The semantics for knowledge then
enforces the intuition that the agent knows neither the state nor the
derandomizer used.%
\footnote{A reviewer of the paper suggested that we instead should
define $(M,s,v) \sat K_i \phi$ if $(M,t,v) \sat \phi$ for all $t \in S$
such that $s \sim_i t$.  This would be appropriate if the agent knew the
derandomizer being used.}

Having the sequence of coin tosses as
part of the input allows us to talk about the probability that $i$'s
algorithm answers yes to the query $\phi$ at a state $s$.  It
is simply $\nu(\{v \mid \Ad_i(\phi,L_i(s),s,v_i)=\YES\})$. To capture this in
the language, we extend the language to allow formulas of the form
$\Pr(\phi) \geq \alpha$, read ``the probability of $\phi$ is at least
$\alpha$''.%
\footnote{We allow $\alpha$ to be an arbitrary real number here.  If we
were concerned with complexity results and having a finitary language,
it would make sense to restrict $\alpha$ to being rational, as is done,
for example, in \cite{FHM}.  None of our results would be affected if we
restrict $\alpha$ in this way.}
The semantics of such formulas is straightforward:
\begin{itemize}
\item[]$(\N,s,v)\sat \Pr(\phi) \ge \alpha$ if $\nu(\{v' \mid  (\N,s,v')
\sat \phi\}) \ge \alpha$.
\end{itemize}
Note that the truth of $\Pr(\phi) \ge \alpha$ at a state $(s,v)$ is
independent of $v$.  Thus, we can abuse notation and write $(\N,s)
\sat \Pr(\phi) \ge \alpha$.  In particular,  $(\N,s) \sat
\Pr(X_i \phi) < \alpha$ (or, equivalently, $(\N,s) \sat \Pr(\neg X_i
\phi) \ge 1-\alpha$) if the  probability of the knowledge algorithm
returning $\YES$ on a query $\phi$ is less than $\alpha$,
given state $s$.

If all the knowledge algorithms used are deterministic, then this
semantics agrees with the semantics given in Section~\ref{s:algknow}.  To
make this precise, note that if $\A$ is deterministic, then
$\Ad(\phi,\ell,v_i) = \Ad(\phi,\ell,v_i')$ for all $v, v'
\in V$.  In this case, we abuse notation and write $\A(\phi,\ell)$.
\pro\label{p:deterministic}
Let $\N=(S,\pi,L_1,\dots,L_n,\Ad_1,\dots,\Ad_n,\nu)$ be a
probabilistic algorithmic knowledge, with $\A_1,\dots,\A_n$
deterministic. Let $M=(S,\pi,L_1,\dots,L_n,\A_1,\dots,\A_n)$. If there
are no occurrences of $\Pr$ in $\phi$ then, for all $s\in S$ and all
$v\in V$, $(\N,s,v)\sat\phi$ if and only if $(M,s)\sat\phi$. 
\epro

Thus, derandomizers are not needed to interpret the $X_i$ operators if
the knowledge algorithms are all deterministic. Moreover, in general,
derandomizers are necessary only to interpret the $\Pr$ and $X_i$ operators. 
\pro\label{p:xfree}
Let $\N=(S,\pi,L_1,\dots,L_n,\Ad_1,\dots,\Ad_n,\nu)$ be a
probabilistic algorithmic knowledge structure and let
$M=(S,\pi,L_1,\dots,L_n,\A'_1,\dots,\A'_n)$ be an algorithmic
knowledge structure, where $\A'_1,\dots,\A'_n$ are arbitrary
deterministic knowledge algorithms. If there are no occurrences of
$X_i$ and $\Pr$ in $\phi$ then, 
for all $s\in S$ and all $v\in V$, 
$(\N,s,v)\sat\phi$ if and only if
$(M,s)\sat\phi$.
\epro

Propositions~\ref{p:deterministic} and \ref{p:xfree} justify the
decision to ``factor out'' the randomization of the knowledge
algorithms into semantic objects that are distinct from the 
states; the semantics of formulas that do not depend on the
randomized choices do not in fact depend on those additional semantic
objects.  
\section{An Example from Security}\label{s:security}

\newcommand{\ADYi}{\A^{\scriptscriptstyle\rm DY}_i}
\newcommand{\ADY}{\A^{\scriptscriptstyle\rm DY}}
\newcommand{\ADYg}{\A^{\scriptscriptstyle{\rm DY}+{\rm rg}(\ng)}}
\newcommand{\ADYgi}{\A^{\scriptscriptstyle{\rm DY}+{\rm rg}(\ng)}_i}
\newcommand{\sees}{\mathsf{has}}
\newcommand{\msg}{\mathsf{m}}
\newcommand{\recv}{\mathsf{recv}}
\newcommand{\tspace}{~~~~~}
\newcommand{\encr}[2]{\{\!\hspace{-.7pt}|#1|\!\hspace{-.7pt}\}_{#2}}
\newcommand{\pl}{\mathsf{p}}
\newcommand{\ky}{\mathsf{k}}
\newcommand{\derivesDY}{\vdash_{\scriptscriptstyle DY}}

As we mentioned in the introduction, an 
important area of
application for algorithmic knowledge is the analysis of cryptographic
protocols. In previous work \cite{r:halpern02e}, we showed how
algorithmic knowledge can be used to model the resource limitations of
an adversary. We briefly review the framework of that paper here.

Participants in a security protocol are viewed as exchanging messages
in the
free algebra generated by a set $\cP$ of plaintexts and a set
$\cK$ of keys, over abstract operations $\cdot$ 
(concatenation) and $\encr{~}{}$ (encryption).
The set $\cM$ of messages is the smallest set that contains $\cK$ and
$\cP$ and is closed under encryption and concatenation, so that if
$\msg_1$ and $\msg_2$ are in $\cM$ and $\ky \in \cK$, then $\msg_1 \cdot
\msg_2$ and $\encr{\msg_1}{\ky}$ are in $\cM$.  We
identify elements of $\cM$
under the equivalence $\encr{\encr{\msg}{\ky}}{\ky^{-1}}=\msg$.  We
make the assumption, standard in the security literature, that
concatenation and encryption have 
enough redundancy to recognize that a term is in fact a concatenation
$\msg_1\cdot\msg_2$ or an encryption $\encr{\msg}{\ky}$. 

\commentout{
Define a ``submessage'' relation $\sqsubseteq$ on $\cM$ as the
smallest relation satisfying the following constraints:
\begin{enumerate}
\item $\msg\sqsubseteq \msg$;
\item if $\msg\sqsubseteq \msg_1$, then $\msg\sqsubseteq \msg_1\cdot \msg_2$;
\item if $\msg\sqsubseteq \msg_2$, then $\msg\sqsubseteq \msg_1\cdot \msg_2$;
\item if $\msg\sqsubseteq \msg_1$, then $\msg\sqsubseteq \encr{\msg_1}{\ky}$.
\end{enumerate}
Intuitively, $\msg_1\sqsubseteq \msg_2$ if $\msg_1$ \emph{could} be
used in the construction of $\msg_2$. For example, if
$\msg=\encr{\msg_1}{\ky}=\encr{\msg_2}{\ky}$, then both
$\msg_1\sqsubseteq \msg$ and $\msg_2\sqsubseteq \msg$.  Therefore, if
we want to establish that $\msg_1\sqsubseteq \msg_2$ for a given
$\msg_1$ and $\msg_2$, then we have to look at all the possible ways
in which $\msg_2$ can be taken apart, either by concatenation or
encryption, to finally decide if $\msg_1$ can be derived from
$\msg_2$.
}
In an  {\em algorithmic 
knowledge 
security structure}, 
some of the agents are participants in the security protocol being
modeled, while other agents are \emph{adversaries} that do not
participate in the protocol, but attempt to subvert it. 
The adversary is viewed as just another agent,
whose local 
state contains
all the
messages it has intercepted, as well as the keys initially known to
the adversary, such as the public keys of all the agents. We use
$\mathit{initkey}(\ell)$ to denote the set of initial keys known by
an agent with local state $\ell$
and write $\recv(\msg) \in \ell$ if $\msg$ is one of the messages
received (or intercepted in the case of the adversary) by an agent with
local state $\ell$.
We assume 
that the language includes
a primitive proposition
$\sees_i(\msg)$ for every message $\msg$, essentially saying that
message $\msg$ is contained within a message that agent $i$ has
received.  
Define the containment relation $\sqsubseteq$ on $\cM$ as the smallest
relation satisfying the following constraints: 
\begin{enumerate}
\item $\msg\sqsubseteq \msg$;
\item if $\msg\sqsubseteq \msg_1$, then $\msg\sqsubseteq \msg_1\cdot \msg_2$;
\item if $\msg\sqsubseteq \msg_2$, then $\msg\sqsubseteq \msg_1\cdot \msg_2$;
\item if $\msg\sqsubseteq \msg_1$, then $\msg\sqsubseteq \encr{\msg_1}{\ky}$.
\end{enumerate}
Formally, $\sees_i(\msg)$ is true at a local state $\ell$ if $\msg
\sqsubseteq \msg'$ for some message $\msg'$ such that $\recv(\msg')
\in \ell$.  

Clearly, the adversary may not explicitly know that he has a given
message if that message is encrypted using a key that the adversary
does not know. To capture these restrictions, Dolev and Yao
\citeyear{r:dolev83} gave a now-standard description of capabilities of
adversaries. Succinctly, a Dolev-Yao adversary can compose messages,
replay them, or decipher them if he knows the right keys, but cannot
otherwise ``crack'' encrypted messages. The Dolev-Yao model can be
formalized by a relation $H\derivesDY \msg$ between a set $H$ of
messages and a message $\msg$.  (Our formalization is equivalent to
many other formalizations of Dolev-Yao in the literature, and is
similar in spirit to that of Paulson \citeyear{r:paulson98}.)
Intuitively, $H\derivesDY \msg$ means that an adversary can
``extract'' message $\msg$ from a set of received messages and keys
$H$, using the allowable operations. The derivation is defined using
the following inference rules:
\[\Rule{\msg\in H}{H\derivesDY \msg} \quad \Rule{H\derivesDY\encr{\msg}{\ky}
\quad H\derivesDY \ky^{-1}}{H\derivesDY \msg} \quad
\Rule{H\derivesDY \msg_1\cdot \msg_2}{H\derivesDY \msg_1} \quad
\Rule{H\derivesDY \msg_1\cdot \msg_2}{H\derivesDY \msg_2},\]
where $\ky^{-1}$ represents the key used to decrypt messages encrypted
with $\ky$. 

We can encode these capabilities via a knowledge algorithm $\ADY$ for
the adversary  as agent $i$.  
Intuitively, the knowledge algorithm $\ADY$ simply implements a search 
for the derivation
of a message $\msg$ from the messages 
that the agent has received and the initial set of keys, using the rules
given above.
The most interesting case in the
definition of $\ADY$ is when the formula is $\sees_i(\msg)$.  To compute
$\ADYi(\sees_i(\msg),\ell,s)$, the algorithm simply checks, for every
message $\msg'$ received by the adversary, whether $\msg$ is a
submessage of $\msg'$, according to the keys that are known to the
adversary
(given by the function $\mathit{keysof}$).
Checking whether $\msg$ is a submessage of $\msg'$ is
performed by a function $\mathit{submsg}$, which can take apart
messages created by concatenation, or decrypt messages as long as the
adversary knows the decryption key.
(The function $\mathit{submsg}$ basically implements the inference
rules for $\derivesDY$.)
$\ADYi(\sees_i(\msg),\ell,s)$ is defined by the following algorithm:
\begin{align*}
 & \text{if $\msg\in\mathit{initkeys}(\ell)$ then return $\YES$}\\
 & \text{$K = \mathit{keysof}(\ell)$}\\
 & \text{for each $\recv(\msg')\in \ell$ do}\\
 & \quad\text{if $\mathit{submsg}(\msg,\msg',K)$ then}\\
 & \qquad\text{return $\YES$}\\
 & \text{return $\DONTK$.}
\end{align*}
Note that the algorithm does not use the input $s$.
Further details can be found in \cite{r:halpern02e}, 
where it is also shown that 
$\ADYi$ is a sound knowledge algorithm that captures the
Dolev-Yao adversary
in the following sense:
\pro\label{p:dolevyao} 
{\rm\cite{r:halpern02e}}
If $M = (S,\pi,L_1,\ldots,L_n,\A_1,\ldots,\A_n)$ is an
algorithmic knowledge security structure with an adversary as agent $i$ and
$\A_i = \ADY_i$, then $(M,s) \sat X_i(\sees_i(\msg))$ if and only if $\{ \msg \mid
\recv(\msg)\in L_i(s)\} \union \mathit{initkeys}(\ell) \derivesDY
\msg$.  Moreover, if $(M,s) \sat X_i(\sees_i(\msg))$ then $(M,s) \sat
\sees_i(\msg)$.
\epro
The Dolev-Yao algorithm is deterministic.  It does not capture, for
example, an adversary who guesses keys in an effort to crack an
encryption.  Assume that the key space consists of finitely many keys,
and let 
$\mathit{guesskeys}(\ng)$ return $\ng$ of these, chosen uniformly at
random.  Let $\ADYgi$ be the result of modifying 
$\ADYi$ to take random guessing into account (the rg stands for
\emph{random guess}),
so that $\ADYgi(\sees_i(\msg),\ell,s)$ is defined by the following algorithm:
\begin{align*}
 &  \text{if $\msg\in\mathit{initkeys}(\ell)$ then return $\YES$}\\
 &  \text{$K = \mathit{keysof}(\ell) \union \mathit{guesskeys}(\ng)$}\\
 &  \text{for each $\recv(\msg')\in\ell$ do}\\
 &  \quad\text{if $\mathit{submsg}(\msg,\msg',K)$ then}\\
 &  \qquad\text{return $\YES$}\\
 &  \text{return $\DONTK$.}
\end{align*}
(As before, the algorithm does not use the input $s$.)
Using $\ADYgi$, the adversary gets to work with whatever
keys he already had available, all the keys he can obtain using the
standard Dolev-Yao algorithm, 
and the additional $r$ randomly chosen keys returned by 
$\mathit{guesskeys}(\ng)$.  

\newcommand{\abs}[1]{\lvert#1\rvert}

Of course,
if the total number of keys is large relative to $\ng$, making
$\ng$ random guesses should not help much.  Our framework lets us make
this precise.

\pro\label{p:probalgk} 
Suppose that $\N = (S,\pi,L_1,\ldots,L_n,\A_1^d,\ldots,\A_n^d,\nu)$ 
is a probabilistic algorithmic knowledge security structure with an adversary
as agent $i$ and that
$\A_i = \ADYgi$.
Let $K$ be the number
of distinct keys used in the messages in the adversary's local state
$\ell$ (i.e., the number of keys used in the messages that the
adversary has intercepted at a state $s$ 
with $L_i(s)=\ell$).
Suppose that $K/\abs{\cK} < 1/2$ 
and that $\nu$ is the uniform distribution on sequences of coin tosses.
If $(\N,s,v) \sat \neg K_i X_i (\sees_i(\msg))$,
then $(\N,s,v) \sat \Pr(X_i (\sees_i(\msg))) < 1 - e^{-2\ng K/\abs{\cK}}$.
Moreover, if 
$(\N,s,v) \sat X_i(\sees_i(\msg))$ then $(\N,s,v) \sat \sees_i(\msg)$.
\epro

Proposition~\ref{p:probalgk}
says that what we expect to be true is in
fact true: random guessing of keys 
is sound, but it
does not help much (at least, if
the number of keys guessed is a small fraction of the total numbers of
keys).  
If it is possible that the adversary does not have algorithmic
knowledge of $\msg$, then the probability that he has algorithmic
knowledge is low.  While this result just formalizes our intuitions,
it does show that the probabilistic algorithmic knowledge framework has
the resources to formalize these intuitions naturally.

\section{Probabilistic Algorithmic Knowledge}\label{s:probalgknow}

While the ``guessing'' extension of the Dolev-Yao algorithm considered
in the previous section is sound, we are often interested in
randomized
knowledge algorithms that may sometimes make mistakes. 
We consider a number of examples in this section, to motivate our approach.

First, suppose  that Bob knows (or believes) that a coin
is either fair or double-headed, and wants to determine which.
He cannot examine the coin, but he can ``test'' it 
by having it tossed and observing the outcome. Let
$\pdhead$ be a proposition 
that is
true if and only if the coin is
double-headed.
Bob uses the following 
$\pdhead$-complete
randomized knowledge algorithm
$\ABob$: when queried about $\pdhead$, the algorithm ``tosses'' the
coin, returning $\YES$ if the coin lands heads and 
$\NO$ if the coin lands tails.  It is not hard to check that 
if 
the coin is double-headed, 
then
$\ABob$ answers $\YES$ with probability $1$ (and
hence $\NO$ with probability $0$); if
the coin is fair, then $\ABob$ answers $\YES$ 
with probability $0.5$ (and hence $\NO$ 
with probability $0.5$ as well).  Thus, if the coin fair, there is a chance
that $\ABob$ will make a mistake, although we can make the probability
of error arbitrarily small by applying the algorithm repeatedly
(alternatively, by increasing the number of coin tosses performed by
the algorithm).

Second, consider a robot navigating, using a probabilistic
sensor. This sensor returns the distance to the wall in front of the
robot, within some tolerance.
For simplicity, suppose that if
the wall is at
distance $m$, then the sensor will return a reading of $m-1$ with
probability $1/4$, a reading of $m$ with probability $1/2$, and a
reading of $m+1$ with probability $1/4$. 
Let $\pwall(m)$ be a proposition true at a state if and only if the
wall is at distance at most $m$ in front of the robot. 
Suppose 
that the robot uses the 
following 
knowledge algorithm $\ARobot$ to answer queries.
Given query $\pwall(m)$, $\ARobot$ observes the sensor.  Suppose that it
reads $r$.
If $r \le m$, the algorithm returns $\YES$, otherwise, it
returns $\NO$. It is not hard to check that 
if the wall is actually at
distance less than or equal to $m$, then $\ARobot$ answers $\YES$ to a
query $\pwall(m)$ with probability $\le 3/4$ (and hence $\NO$ with
probability $\ge 1/4$). If the wall is actually at distance 
greater than
$m$, then $\ARobot$ answers $\YES$ 
with probability $\le 1/4$ (and hence $\NO$ with a probability $\ge 1/4$).
There are two ways of modeling this situation.  The first (which is what
we are implicitly doing) is to make the reading of the sensor part of
the knowledge algorithm.  This means that the actual reading is not part
of the agent's local state, and that the output of the knowledge
algorithm depends on the global state.
The alternative would have been to model the process of reading the
sensor in the agent's local state.  In that case, the output of the
knowledge algorithm would depend only on the agent's local state.
There is a tradeoff here.  While on the one hand it is useful to have the
flexibility of allowing the knowledge algorithm to depend on the global
state, ultimately, we do not want the knowledge algorithm to use
information in the global state that is not available to the agent.
For example, we would not want the knowledge algorithm's answer to
depend on the actual distance to the wall (beyond the extent to which
the sensor reading depends on the actual distance). 
It is up to the modeler to ensure that the knowledge algorithm is
appropriate. A poor model will lead to poor results.

Finally,
suppose that Alice has in her local state
a number $n > 2$.
Let
$\pprime$ be a proposition true 
at state $s$
if and only if the number $n$ in Alice's local state is prime.
Clearly, Alice
either (implicitly) knows $\pprime$ or knows
$\neg\pprime$. However, this is implicit knowledge.
Suppose that Alice uses Rabin's \citeyear{Rabin80} primality-testing
algorithm to test if $n$ is prime.
That algorithm uses a (polynomial-time computable) predicate
$P(n,a)$ with the following properties, for
a natural number $n$ and $1\leq a\leq n-1$:
\begin{enumerate}
\item $P(n,a)\in\{0,1\}$;
\item if $n$ is composite, $P(n,a)=1$ for at least $n/2$
choices of $a$;
\item if $n$ is prime, $P(n,a)=0$ for all $a$. 
\end{enumerate}
Thus, Alice uses the following 
randomized knowledge algorithm $\AAlice$: 
when queried about $\pprime$, the algorithm
picks a number $a$ at random
between $0$ and the number $n$ in Alice's local state; if $P(n,a)=1$, it
says $\NO$ 
and if $P(n,a)=0$, it says
$\YES$. 
(It is irrelevant for our purposes what the algorithm does on other
queries.) 
It is not hard to 
check that $\AAlice$ has the following properties.
If the number $n$ 
in Alice's local state is prime, then $\AAlice$ answers
$\YES$ to a query $\pprime$ with probability $1$ (and hence $\NO$ to
the same query with probability $0$).  
If $n$ is composite, 
$\AAlice$ answers $\YES$ to a query
$\pprime$ with probability $\leq 1/2$ and $\NO$ with
probability $\geq 1/2$.
Thus, if $n$ is composite, there is a chance that $\AAlice$ will make a
mistake, although we can make the probability of error 
arbitrarily small by applying the algorithm repeatedly.
While this problem seems similar to the double-headed coin example
above, note that we  have only \emph{bounds} on the probabilities
here. The actual probabilities corresponding to a particular number
$n$ depend on various number theoretic properties of that number. We
return to this issue in Section~\ref{s:evidencealg}.

Randomized knowledge algorithms like those in the examples above are
quite common in the literature.  They are not sound, but are ``almost
sound''.  The question is what we can learn from such an ``almost
sound'' algorithm.  
Returning to the first example, we know the probability that $\ABob$
says $\YES$ (to the query $\pdhead$) given that the coin is
double-headed; what we are interested in is the probability that the
coin is double-headed given that $\ABob$ says $\YES$.  (Of course, the
coin is either double-headed or not.  However, if Bob has to make
decisions based on whether the coin is double-headed, it seems
reasonable for him to ascribe a subjective probability to the coin
being double-headed.  It is this subjective probability that we are
referring to here.)

Taking ``dh'' to represent the event ``the coin is double-headed''
(thus, the proposition $\pdhead$ is true at exactly the states in dh),
by
Bayes' rule, 
\[ \Pr(\mbox{dh} ~|~\mbox{$\ABob$ says \YES}) = 
  \frac{ \Pr (\mbox{$\ABob$ says \YES}~|~\mbox{dh})
\Pr(\mbox{dh})}
         { \Pr (\mbox{$\ABob$ says \YES})}.\]
The only piece of information in this equation that we have is
$\Pr(\mbox{$\ABob$ says \YES} ~|~ \mbox{dh})$.
If we had $\Pr(\mbox{dh})$, we could derive
$\Pr(\mbox{$\ABob$ says \YES})$. However, we do not have that
information, since we did not assume a probability distribution on the
choice of coin.  
Although we do not have the information needed to compute
$\Pr(\mbox{dh} ~|~\mbox{$\ABob$ says \YES})$,
there is still a strong intuition that if $X_i\pdhead$ holds, this tells 
us something about whether the coin is double-headed. How can 
this be formalized?

\subsection{Evidence}\label{s:evidence}

Intuitively, the fact that
$X_i\phi$ holds provides 
``evidence'' that $\phi$ holds. But
what is evidence? 
There are a number of definitions in the literature.
They all essentially give a way to assign a ``weight'' to different
hypotheses based on an observation; 
they differ in exactly how they assign the weight (see 
\cite{r:kyburg83} for a survey).  
Some of these approaches make sense only if there is a probability
distribution on the hypotheses.  Since this is typically not the case
in the applications of interest to us
(for example, in the primality example, we do not want to assume a
probability on the input $n$),
we use a definition of evidence given by Shafer
\citeyear{Shafer82} and Walley \citeyear{Walley87}, which does not
presume a probability on hypotheses.
We start with a set $\cH$ of hypotheses, which we take to be mutually
exclusive and exhaustive; 
thus, exactly one hypothesis holds at any given time. For the examples
of this paper, the hypotheses of interest have the form
$\cH=\{h_0,\neg h_0\}$, where the hypothesis $\neg h_0$ is
the negation of hypothesis $h_0$. Intuitively, this is because we want
to reason about the evidence associated with a formula or its negation
(see Section~\ref{s:reliable}).
For example, if $h_0$ is ``the coin is double-headed'', then
$\neg h_0$ is ``the coin is not double-headed'' (and thus, if
there are only two kinds of coins, double-headed and fair, then
$\neg h_0$ is ``the coin is fair'').
We are
given a set $\cO$ of \emph{observations}, which can be understood as
outcomes of experiments that we can make. 
Assume that for each
hypothesis $h\in\cH$ there is a probability space $(\cO,2^{\cO},\mu_h)$.
Intuitively, $\mu_h(\ob)$  is the probability of $\ob$
given that hypothesis $h$ holds. 
While this looks like a conditional probability, 
notice that it does not require a probability on $\cH$.
Taking $\Delta(\cO)$ to denote the set of probability measures on $\cO$,
define an \emph{evidence space} to be a tuple
$\cE=(\cH,\cO,\cF)$, where $\cH$, $\cO$, and
$\cF: \cH \rightarrow \Delta(\cO)$.  Thus, $\cF$ associates with each
hypothesis a probability on observations (intuitively, the probability 
that various observations are true given that the hypothesis holds).  We
often denote $\cF(h)$ as $\mu_h$.
For an evidence space $\cE$, 
the weight that the
observation $\ob$ lends to hypothesis $h\in\cH$, written
$w_{\cE}(\ob,h)$,
is
\begin{equation}\label{e:weight}
 w_{\cE}(\ob,h) =
\frac{\mu_h(\ob)}{\sum_{h'\in\cH}\mu_{h'}(\ob)}.
\end{equation}
Equation~\eqref{e:weight} does not define a weight $w_{\cE}$ for an
observation $\ob$ such that 
$\sum_{h\in\cH}\mu_h(\ob)=0$.
Intuitively, this means that the observation $\ob$ is impossible.
In the literature on confirmation theory it is typically assumed 
that this case never 
arises.  More precisely, it is assumed that 
all observations are possible, so that  for every observation $\ob$,
there is an hypothesis $h$ such that $\mu_h(\ob)>0$. In our case,
making this assumption is unnatural.
We want to view the answers given by
knowledge algorithms as observations, and it seems perfectly
reasonable to have a knowledge algorithm that never returns $\NO$,
for instance. 
As we shall see (Proposition~\ref{p:defined}), the fact that the weight
of evidence is 
undefined in the case that
$\sum_{h\in\cH}\mu_h(\ob) = 0$
is not a problem in our intended application, thanks 
to
our assumption that $\nu$ does not assign zero probability to the
nonempty sets of sequences of coin tosses that determine  the result
of the knowledge algorithm. 

Observe that the measure $w_{\cE}$ always lies between 0 and 1, with 1
indicating that the full weight of the evidence 
of observation $\ob$ is provided to 
the hypothesis.
While the weight of evidence $w_{\cE}$ looks like a probability measure
(for instance, for each fixed observation $\ob$ for which
$\sum_{h\in\cH}\mu_h(\ob)>0$, the sum $\sum_{h\in\cH}w_{\cE}(\ob,h)$
is $1$), one should not interpret it
as a probability measure. It is simply a way to assign
a weight to hypotheses given observations.  
It is possible to interpret the weight function $w$ as a prescription
for how to update a prior probability on the hypotheses into a
posterior probability on those hypotheses, after having considered the
observations made. We do not focus on these aspects here; see
\cite{r:halpern03b} for more details.

For the double-headed coin example, the set $\cH$ of hypotheses is
$\{\pdhead,\neg\pdhead\}$. The observations $\cO$ are simply the
possible outputs of the knowledge algorithm $\ABob$ on the formula
$\pdhead$, namely, $\{\YES,\NO\}$.  From the discussion following the
description of the example, it follows that $\mu_\pdhead(\YES)=1$ and
$\mu_\pdhead(\NO)=0$, since the algorithm always says $\YES$
when the coin is double-headed.  
Similarly, $\mu_{\neg \pdhead}(\YES)$ is the probability that the
algorithm says $\YES$ if 
the coin is not double-headed.  
By assumption, the coin is fair if it is not double-headed, so
$\mu_{\neg\pdhead}(\YES) = 1/2$ and $\mu_{\neg\pdhead}(\NO) =1/2$.  
Define $\cF(\pdhead) = \mu_{\pdhead}$ and $\cF(\neg \pdhead) =
\mu_{\neg\pdhead}$, and   
let
\[\cE=(\{\pdhead,\neg\pdhead\},\{\YES,\NO\},\cF).\]
It is easy to check that $w_{\cE}(\YES,\pdhead)=2/3$ and
$w_{\cE}(\YES,\neg\pdhead)=1/3$.  Intuitively, a $\YES$
answer to the query $\pdhead$ provides more evidence for the
hypothesis $\pdhead$ than the hypothesis $\neg \pdhead$.  Similarly,
$w(\NO,\pdhead)=0$ and $w(\NO,\neg\pdhead)=1$.  Thus, an output of
$\NO$ to the query $\pdhead$ indicates that the hypothesis
$\neg\pdhead$ must hold.

This approach, however, is not quite sufficient to deal with the
sensor example 
because, in that example,
the
probability of an observation does not
depend solely on whether the hypothesis is true or false.
The
probability of the algorithm answering $\YES$ to a query $\pwall(10)$ 
when $\pwall(10)$ is true depends on the actual distance 
$m$ 
to the wall:
\begin{itemize}
\item if $m\leq 9$, then $\mu_{\pwall(10)}(\YES)=1$ (and thus
$\mu_{\pwall(10)}(\NO)=0$);
\item if $m=10$, then $\mu_{\pwall(10)}(\YES)=3/4$ (and thus
$\mu_{\pwall(10)}(\NO)=1/4$).
\end{itemize}
Similarly, the probability of the algorithm answering $\YES$ to a
query $\pwall(10)$ 
in a state where $\neg\pwall(10)$ holds depends on $m$ in the following way:
\begin{itemize}
\item if $m=11$, then $\mu_{\neg\pwall(10)}(\YES)=1/4$;
\item if $m\ge 12$, then $\mu_{\neg\pwall(10)}(\YES)=0$.
\end{itemize}
It does not seem possible to capture 
this information using the type of evidence space defined above.   In 
particular,  we do not have a 
single probability measure over the observations given a particular
hypothesis. 
One reasonable way of 
capturing the information is to associate a \emph{set} of
probability measures on observations with each hypothesis;
intuitively, these represent the possible probabilities on the
observations, depending on the actual state.

To make this precise,
define a \emph{generalized evidence space} to be a tuple
$\cE=(\cH,\cO,\cF)$, where now $\cF: \cH \rightarrow 2^{\Delta(\cO)}$.
We require $\cF(h) \ne \emptyset$ for at least one $h \in \cH$.
What is the most appropriate way to define weight of evidence given sets
of probability measures?  
As a first step, 
consider the set of all possible weights of
evidence that are obtained by taking any combination of probability
measures, one from each set  $\cF(h)$ (provided that $\cF(h) \ne \emptyset$).
This gives us a range of possible weights of evidence. 
We can then define \emph{upper} and \emph{lower} weights of
evidence, determined by the maximum and minimum values in the range,
somewhat analogous to the notions of upper and lower probability
\cite{Hal31}.  (Given a set $\cP$ of probability measures, the
\emph{lower probability} of a set $U$ is $\inf_{\mu \in \cP} \mu(U)$;
its \emph{upper probability} is $\sup_{\mu \in \cP} \mu(U)$.)
Let
\[ 
\cW_{\cE}(\ob,h)=\left\{\frac{\mu_h(\ob)}{\sum_{h'\in\cH,\cF(h') \ne
\emptyset}\mu_{h'}(\ob)} ~\Bigm|~ \mu_{h}\in\cF(h),\mu_{h'}\in\cF(h'), \, \sum_{\substack{h'\in\cH\\\cF(h') \ne
\emptyset}}\mu_{h'}(\ob) \ne 0 \right\}.\] 
Thus, $\cW_{\cE}(\ob,h)$ is the set of possible weights of evidence
for the hypothesis $h$ given by $\ob$.  Define the \emph{lower weight
of evidence function} $\wlow_{\cE}$ by taking $\wlow_{\cE}(\ob,h)=\inf
\cW_{\cE}(\ob,h)$; similarly, define the \emph{upper weight of
evidence function} $\wup_{\cE}$ by taking $\wup_{\cE}(\ob,h)=\sup
\cW_{\cE}(\ob,h)$.  If $\cW_{\cE}(\ob,h) =
\emptyset$, which will happen either if $\cF(h) = \emptyset$ or if
$\sum_{h'\in\cH,\cF(h') \ne \emptyset}\mu_{h'}(\ob) = 0$ for all
choices of $\mu_{h'} \in \cF(h')$ for $\cF(h') \ne \emptyset$, then we
define $\wlow_{\cE}(\ob,h)=\wup_{\cE}(\ob,h) = 0$.  We show in
Proposition~\ref{p:defined} that, in the special case where $\cF(h)$
is a singleton for all $h$ (which has been the focus of all previous
work in the literature), $\cW_{\cE}(\ob,h)$ is a singleton under our
assumptions.  In particular, the denominator is not 0 in this case.
Of course, if $\cF(h)= \{\mu_h\}$ for all hypotheses $h\in\cH$, then 
$\wlow_{\cE} = \wup_{\cE} = w_{\cE}$.

Lower and upper evidence can be used to model the
examples at the beginning of this section. 
In the sensor example, 
with $\cH=\{\pwall(10),\neg\pwall(10)\}$, 
there are two probability measures associated with
the hypothesis $\pwall(10)$, namely, 
\begin{align*}
\mu_{\pwall(10),\le 9}(\YES) & = 1\\
\mu_{\pwall(10),=10}(\YES) & =3/4;
\end{align*}
similarly, there are two probability measures associated with the
hypothesis $\neg\pwall(10)$, namely
\begin{align*}
\mu_{\neg\pwall(10),=11}(\YES)&=1/4\\
\mu_{\neg\pwall(10),\ge 12}(\YES)&=0.
\end{align*}
Let $\cE$ be the corresponding eneralized evidence space. 
It is easy to check that
\[\cW_\cE(\YES,\pwall(10))=\{4/5,1,3/4\},\] and thus 
\[\text{$\wlow_\cE(\YES,\pwall(10)) =3/4$ and $\wup_\cE(\YES,\pwall(10)) =1$.}\]
Indeed,
using $\mu_{\pwall(10),\le 9}$ and $\mu_{\neg\pwall(10),=11}$ 
gives $4/5$; using $\mu_{\pwall(10),=10}$ and $\mu_{\neg\pwall(10),=11}$ 
gives $3/4$; and using either $\mu_{\pwall(10),\le 9}$ or 
$\mu_{\pwall(10),=10}$ with $\mu_{\neg\pwall(10),\ge 12}$ gives 1.
Similarly, 
\[\cW_\cE(\YES,\neg\pwall(10))=\{1/5,1/4,0\},\] and thus 
\[\text{$\wlow_\cE(\YES,\neg\pwall(10)) = 0$ and $\wup_\cE(\YES,\neg\pwall(10)) =1/4$}.\]
In particular,
if the algorithm
answers $\YES$ to a query $\pwall(10)$, the evidence supports the
hypothesis that the wall is indeed at a distance less than 10 from the 
robot.

The primality example can be dealt with in the same way.
Take $\cH=\{\pprime,\neg\pprime\}$. There is a single probability
measure $\mu_{\pprime}$ associated with the hypothesis $\pprime$,
namely $\mu_{\pprime}(\YES)=1$; intuitively, if the number is prime,
the knowledge algorithm always returns the right answer. In contrast,
there are a number of different probability measures
$\mu_{\neg\pprime,n}$ associated with the hypothesis $\neg\pprime$,
one per composite number $n$, where we take 
$\mu_{\neg\pprime,n}(\YES)$ to be the probability that the algorithm
says $\YES$ when the composite number $n$ is in Alice's local
state. 
Note that this probability is 1 minus the fraction of ``witnesses'' $a <
n$ such that $P(n,a) = 1$.
The fraction of witnesses
depends on number-theoretic properties of $n$, and thus may be
different for different choices of composite numbers $n$.
Moreover, Alice
is unlikely to know the actual probability $\mu_{\neg\pprime,n}$.  
As we mentioned above, it has been shown that $\mu_{\neg\pprime,n} \le
1/2$ for all composite $n$, but Alice may not know any more than this.
Nevertheless, for now, we assume that Alice is an ``ideal'' agent who
knows the set
$\{\mu_{\neg \pprime,n}\mid \text{$n$ is composite}\}$.
(Indeed, in the standard Kripke structure framework for knowledge, it is
impossible to assume anything else!)
We consider how to model the set of
probabilities 
used by 
a ``less-than-ideal'' agent in Section~\ref{s:evidencealg}.
Let $\cE$ be the corresponding generalized evidence
space.  Then 
\[ \cW_\cE(\YES,\pprime)=\{
1/(1+\mu_{\neg\pprime,n}(\YES))\mid \text{$n$ composite}\}.\]
Since $\mu_{\neg\pprime,n}(\YES)\le 1/2$ for all composite $n$,
it follows that
$\wlow_\cE(\YES,\pprime)\ge 2/3$. Similarly,
\[\cW_\cE(\YES,\neg\pprime) =
\{\mu_{\neg \pprime,n}(\YES)/(\mu_{\neg \pprime,n}(\YES)+1)\mid\text{$n$
composite}\}.\]
Since $\mu_{\neg\pprime,n}(\YES)\le 1/2$ for all composite
$n$, we have that
$\wup_\cE(\YES,\neg\pprime)\le 1/3$. Therefore, if the
algorithm answers $\YES$ to a query $\pprime$, the evidence supports
the hypothesis that the number 
is indeed 
prime. 

Note that, in modeling this example, we have assumed that the number $n$
is {\em not\/} in Alice's local state and that Alice 
knows the fraction of witnesses $a$ for each composite number $n$.
This means that the same set of
probabilities used by Alice for all choices of $n$ (since the
set of probabilities used depends only on Alice's local state), and
is determined by the set of possible fraction of elements $< n$ that are
witnesses, for each composite number $n$.
Assuming that $n$ is in Alice's local state (which is actually quite a
reasonable assumption!) and that Alice does not know the fraction of
numbers less than $n$ that are witnesses adds new subtleties; we consider
them in Section~\ref{s:evidencealg}.

\subsection{Evidence for Randomized Knowledge Algorithms}\label{s:evidencealg} 

We are now ready to discuss randomized knowledge algorithms. 
What does a $\YES$ answer to a query $\phi$ given by an ``almost sound''
knowledge algorithm tell us about $\phi$? As the discussion in
Section~\ref{s:evidence} indicates, a $\YES$ answer to a query $\phi$
provides evidence for the hypotheses $\phi$ and $\neg\phi$. This can be
made precise by associating an evidence space with every state of the
model to capture the evidence provided by the knowledge algorithm.
To simplify the presentation, we restrict our attention to knowledge
algorithms that are $\phi$-complete. (While it is possible to deal
with general knowledge algorithms that also can return $\DONTK$ 
using
these techniques, we already saw that the logic does not let us
distinguish between a knowledge algorithm returning $\NO$ and a
knowledge algorithm returning $\DONTK$; they both result in lack of
algorithmic knowledge.  In the next section, where we define the
notion of a reliable knowledge algorithm, reliability will be
characterized in terms of algorithmic knowledge, and thus the
definition will not distinguish between a knowledge algorithm
returning $\NO$ or $\DONTK$. In order to establish a link between the
notion of reliability and evidence, it is convenient to either
consider $\phi$-complete algorithms, or somehow identify the answers
$\NO$ and $\DONTK$. We choose the former.)  Note that the knowledge
algorithms described in the examples at the beginning of this section
are all complete for their respective hypotheses. We 
further
assume that the truth of $\phi$ depends only on the state, and not on
coin tosses, that is, $\phi$ does not contain occurrences of the $X_i$
operator.

Our goal is to associate, with every local state $\ell$ of agent $i$
in $\N$ an evidence space over the hypotheses $\{\phi,\neg\phi\}$ and
the observations $\{\YES,\NO,\DONTK\}$.
Let $S_\ell=\{s\mid L_i(s)=\ell\}$ be the set of states where
agent $i$ has local state $\ell$. 
At every state $s$ of $S_\ell$, let 
$\mu_{s,\phi}(\ob)=\nu(\{v' \mid
\Ad_i(\phi,\ell,s,v'_i)=\ob\})$.  
Intuitively, $\mu_{s,\phi}$ gives the probability of observing $\YES$ and
$\NO$ in state $s$.
Let $S_{\ell,\phi}=\{s\in S_\ell\mid (\N,s)\sat\phi\}$ and let
$S_{\ell,\neg\phi}=\{s\in S_\ell\mid (\N,s)\sat\neg\phi\}$. 
(Recall that $\phi$ depends only on the state, and not on 
the outcome of the 
coin tosses.) 
Define $\cF_{\ell,\phi}(\phi) = \{\mu_{s,\phi} \mid s \in S_{\ell,\phi}\}$ and 
$\cF_{\ell, \phi}(\neg \phi) = \{\mu_{s,\phi} \mid s \in S_{\ell,\neg
\phi}\}$; then the 
evidence space is
\[\cE_{\A_i,\phi,\ell} = (\{\phi,\neg\phi\},\{\YES,\NO,\DONTK\},
\cF_{\ell,\phi}).\]
(We omit the $\DONTK$ from the set of possible observation if the
knowledge algorithm is $\phi$-complete, as is the case in the three
examples given at the beginning of this section.)
Since the agent does not know which state $s \in S_\ell$ is the true
state, he must consider all the probabilities in $\cF_{\ell,\phi}(\phi)$ and
$\cF_{\ell,\phi}(\neg \phi)$ in his evidence space.

We can now make precise the claim at which we have been hinting
throughout the paper.  Under our assumptions, for all evidence spaces 
of the form $\cE_{\A_i,\phi,\ell}$ that arise in this construction, and
all observations $\ob$ that can be made in local state $\ell$, there
must be some expression in $\cW_{\cE_{\A_i,\phi,\ell}}(\ob,h)$ with a
nonzero denominator.   Intuitively, this is because if $\ob$ is observed
at some state $s$ such that $L_i(s) = \ell$, our assumptions ensure that
$\mu_{s,\phi}(\ob)>0$. 
In other words, observing $\ob$ means that the probability of
observing $\ob$ must be greater than 0.

\pro\label{p:defined}
For all probabilistic algorithmic knowledge structures $N$,
agents $i$, formulas $\phi$, and local states $\ell$ of agent $i$ that
arise in $N$, if $\ob$ is a possible output of $i$'s knowledge algorithm 
$\Ad_i$ in local state $\ell$ on input $\phi$, then there exists a
probability measure $\mu \in \cF_{\ell,\phi}(\phi) \union 
\cF_{\ell,\phi}(\neg \phi)$ such that $\mu(\ob) > 0$.
\epro
In particular, it follows from Proposition~\ref{p:defined}
that, under our assumptions, the evidence function is always defined in
the special case where  $\cF_{\ell,\phi}(h)$ is a singleton for all hypotheses $h$.

To be able to talk about evidence within the logic, we introduce
operators to capture the lower and upper evidence provided by the
knowledge algorithm of agent $i$, $\lEv_i(\phi)$ and $\uEv_i(\phi)$,
read 
``$i$'s lower (resp., upper) weight of evidence for $\phi$'', with
semantics defined as follows:
\begin{itemize}
\item[] $(\N,s,v)\sat \lEv_i(\phi)\geq\alpha$ if
$\wlow_{\cE_{\A_i,\phi,L_i(s)}}(\Ad_i(\phi,L_i(s),s,v_i),\phi)\geq\alpha$
\item[] $(\N,s,v)\sat \uEv_i(\phi)\geq\alpha$ if
$\wup_{\cE_{\A_i,\phi,L_i(s)}}(\Ad_i(\phi,L_i(s),s,v_i),\phi)\geq\alpha$.
\end{itemize}
We similarly define $(\N,s,v)\sat\lEv_i(\phi)\leq\alpha$,
$(\N,s,v)\sat\uEv_i(\phi)\leq\alpha$, 
$(\N,s,v)\sat\lEv_i(\phi)=\alpha$, and
$(\N,s,v)\sat\uEv_i(\phi)=\alpha$.
By Proposition~\ref{p:defined},  these formulas are all well defined.

This definition of evidence has a number of interesting
properties. For instance, 
obtaining
full evidence in support of a formula $\phi$ 
essentially corresponds to establishing the truth of $\phi$.
\pro\label{p:evidence-knowledge}
For all probabilistic algorithmic knowledge structures $\N$, 
we have $$N \sat \lEv_i(\phi)=1\rimp \phi.$$
\epro

Suppose that we now apply the recipe above the derive the evidence
spaces for the three examples at the beginning of this section.
For the
double-headed coin example, consider a structure $N$ with two states
$s_1$ and $s_2$, where the coin is double-headed at state $s_1$ and
fair at state $s_2$, so that $(N,s_1,v)\sat\pdhead$ and
$(N,s_2,v)\sat\neg\pdhead$. 
Since Bob does not know whether the coin is fair or double-headed, it
seems reasonable to
assume that Bob has the same local state
$\ell_0$ at both states. Thus, $S_{\ell_0}=\{s_1,s_2\}$,
$S_{\ell_0,\pdhead}=\{s_1\}$, and $S_{\ell_0,\neg\pdhead}\{s_2\}$. 
Since we are interested only in the query $\pdhead$ and there is only
one local state, we can consider the single evidence space
\[\cE=(\{\pdhead,\neg\pdhead\},\{\YES,\NO\},\cF_{\pdhead}),\]
where 
\begin{align*}
\cF_{\pdhead}(\pdhead) & = \{\mu_{s_1}\}\\
\cF_{\pdhead}(\neg \pdhead) & = \{\mu_{s_2}\}\\[1ex]
\mu_{s_1}(\YES) & =1\\
\mu_{s_2}(\YES) & =1/2.
\end{align*}
We can check that, for all states $(s,v)$ where
$\ABob(\pdhead,\ell_0,s,v_{\mathrm{Bob}})=\YES$, 
$(N,s,v)\sat\lEv(\pdhead)=2/3$ and $(N,s,v)\sat\uEv(\pdhead)=2/3$,
while at all states $(s,v)$ where\break 
$\ABob(\pdhead,\ell_0,s,v_{\mathrm{Bob}})=\NO$, 
$(N,s,v)\sat\lEv(\pdhead)=0$ and $(N,s,v)\sat\uEv(\pdhead)=0$. In
other words, the algorithm answering $\YES$ provides evidence for the
coin being double-headed, while the algorithm answering $\NO$
essentially says that the coin is fair. 

For the probabilistic sensor example, consider a structure $N$ with
states $s_m$ ($m\ge 1$), where the wall at state $s_m$ is at distance
$m$ from the robot. Suppose 
that
we are interested in the hypotheses
$\pwall(10)$ and $\neg\pwall(10)$, so that $(N,s_m,v)\sat\pwall(10)$
if and only $m\le 10$. The local state of the robot is the
same at every state, say $\ell_0$. Thus, $S_{\ell_0}=\{s_m\mid m\ge
1\}$, $S_{\ell_0,\pwall(10)}=\{s_m\mid 1\le m\le 10\}$, and
$S_{\ell_0,\neg\pwall(10)}=\{s_m\mid m\ge 11\}$. 
Again, since there is only one local state and we are interested in only
one query ($\pwall(10)$ we can consider the single evidence space
\[\cE=(\{\pwall(10),\neg\pwall(10)\},\{\YES,\NO\}, \cF_{\pwall(10)}),\]
where 
\begin{align*}
\cF_{\pwall(10)}(\pwall(10)) & = \{\mu_m \mid 1 \le m \le 10\}\\
\cF_{\pwall(10)}(\neg \pwall(10)) & = \{\mu_{m}\mid m\ge 11\})\\
\mu_m(\YES) & = \begin{cases}
   1 & \text{if $m\le 9$}\\
   3/4 & \text{if $m=10$}\\
   1/4 & \text{if $m=11$}\\
   0 & \text{if $m\ge 12$.}
              \end{cases}
\end{align*}
It is straightforward to compute
that, for all states $(s,v)$ where
$\ARobot(\pwall(10),\ell_0,s,v_{\mathrm{Robot}})$ $=\YES$, 
$(N,s,v)\sat\lEv(\pwall(10))\ge 3/4$ and
$(N,s,v)\sat\uEv(\pwall(10))\le 1$, while at all states $(s,v)$ where 
$\ARobot(\pwall(10),\ell_0,s,v_{\mathrm{Robot}})=\NO$, 
$(N,s,v)\sat\uEv(\pwall(10))\le 1/4$ and
$(N,s,v)\sat\lEv(\pwall(10))\ge 0$. In other words, the algorithm
answering $\YES$ provides evidence for the wall being at distance at
most 10, while the algorithm answering $\NO$ provides evidence for the
wall being further away.

Finally, we consider the primality example. 
Earlier we discussed this example under the assumption that the number
$n$ was not part of Alice's local state.  Under this assumption, it seems
reasonable to assume that there is only one local state, call it $\ell$,
and that we can identify the global state with the number $n$.  
Thus, $S_{\ell,\pprime} = \{n \mid n \mbox{ is prime}\}$ and
$S_{\ell,\neg \pprime} = \{n \mid n \mbox{ is not prime}\}$.
Define $\cF_{\pprime}(\pprime) = \{\mu_{\pprime}\}$, where
$\mu_\pprime(\YES)  = 
1$, while $\cF_{\pprime}(\neg \prime) = \{\mu_{n}\mid n \mbox{ is not
prime}\}$, where $\mu_n(\YES)$ is the fraction of numbers $a <n$ such
that $P(n,a) = 0$.  

What should we do if Alice knows the input (so that $n$ is part of
the local state)?  In that case, it seems that the obvious thing to do
is to again have one state denoted $n$ for every number $n$, but since
$n$ is now part of the local state, we can take $S_n = \{n\}$.
But modeling things this way also points out a problem.  With this state
space, since the agent considers only one state possible in each local
state, it is easy to check that $(N,s,v) \sat \lEv(\pprime) = 1$ if
$s\in S_n$ with $n$ prime, and $(N,s,v) \sat \lEv(\neg \pprime) = 1$
if $s\in S_n$ with $n$ not
prime.  The knowledge algorithm is not needed here.  Since the basic
framework implicitly assumes that agents are logically omniscient, Alice
knows whether or not $n$ is prime.

To deal with this, we need to  model agents that are not logically
omniscient.  Intuitively, we would like to model Alice's subjective view
of the number.  If she does not know whether the number $n$ is prime,
she must consider 
possible a world where $n$ is prime and a world where $n$ is not prime.
We should allow her to consider possible a world where $n$ is prime, and
another world where $n$ is not prime.  of course, if $n$ is in fact
prime, then the world where $n$ is not prime is what Hintikka
\citeyear{Hi2} has called an \emph{impossible possible worlds},
one where the usual laws of arithmetic do not hold.
Similarly, 
since Alice does not know how likely the
knowledge algorithm is to return ``Yes'' if $n$ is composite
(i.e., how many witnesses $a$ there are such that $P(n,a)=0$),
then we should allow her to consider possible the impossible worlds
where the number of witnesses is $k$ for each $k > n/2$.  (We
restrict to $k > n/2$ to model the fact that Alice does know that there 
are at least $n/2$ witnesses if $n$ is composite.)
Thus,
consider the structure $N$ with states $s_{n,\pprime}$ and
$s_{n,\neg\pprime,k}$ (for $n\ge 2$, $n/2 < k \le n$).
Intuitively, $s_{n,\neg\pprime,k}$ is the state where there are $k$ witnesses.
(Clearly, if there is more information about the number of witnesses,
then the set of states should be modified appropriately.)
At states $s_{n,\pprime}$ and 
$s_{n,\neg\pprime,\alpha}$, 
Alice has the same local state, which we call $\ell_n$ (since we
assume that $n$ is stored in her local state); however
$(N,s_{n,\pprime},v)\sat\pprime$, while
$(N,s_{n,\neg\pprime,k},v)\sat\neg\pprime$. 
For a local state
$\ell_n$, define 
$S_{\ell_n,\pprime}=\{s_{n,\pprime}\}$, and
$S_{\ell_n,\neg\pprime}=\{s_{n,\neg\pprime,k}\mid n/2 < k \le n\}$,
and let $S_{\ell_n} = S_{\ell_n,\pprime} \union S_{\ell_n,\neg\pprime}$.
In this model, the evidence space at local state $\ell_n$ is therefore
\[\cE_n=(\{\pprime,\neg\pprime\},\{\YES,\NO\}, \cF_{\ell_n,\pprime}),\]
where 
\begin{align*}
\cF_{\ell,n}(\pprime) & = \{\mu_{n,\pprime}\}\\
\cF_{\ell,n}(\neg \pprime) & =  
\{\mu_{n,\neg\pprime,k}\mid n/2 < k \le n\}\\
\mu_{n,\pprime}(\YES) & =1\\
\mu_{n,\neg\pprime,k}(\YES) & =1-k/n.
\end{align*}
Using impossible possible worlds in this way gives us just the answers
we expect.  
We can check that, for all states $(s,v)$ where
$\AAlice(\pprime,s_{\mathrm{Alice}},s,v)=\YES$,
$(N,s,v)\sat\lEv(\pprime)\ge 2/3$, while at all states $(s,v)$ where
$\AAlice(\pprime,s_{\mathrm{Alice}},s,v)=\NO$,
$(N,s,v)\sat\uEv(\pprime)=0$. In other words, the algorithm returning
$\YES$ to the query whether the number in Alice's local state is
prime provides evidence for the number being prime, while the
algorithm returning $\NO$ essentially says that the number is
composite.

\subsection{Reliable Randomized Knowledge Algorithms}\label{s:reliable}

As we saw in the previous section,  a $\YES$ answer to a query $\phi$
given by an ``almost sound'' knowledge algorithm provides evidence for
$\phi$. We now examine the extent to which we can characterize
the evidence provided by a randomized knowledge algorithm.
To make this precise, we need to first characterize how reliable the
knowledge algorithm is.
(In this section, for simplicity, we assume that we are dealing with complete
algorithms, which always answer either $\YES$ or $\NO$. Intuitively,
this is because reliability, as we will soon see, talks about the
probability of a knowledge algorithm answering $\YES$ or anything but
$\YES$. Completeness ensures that there is a single observation that
can be interpreted as not-$\YES$; this lets us relate reliability
to our notion of evidence in Propositions~\ref{p:properties} and
\ref{p:negproperties}. Allowing knowledge algorithms to return both
$\NO$ and $\DONTK$ would require us to talk about the evidence 
provided
by the disjunction $\NO$-or-$\DONTK$ of the observations, a topic
beyond the scope of this paper.)  
A randomized knowledge algorithm
$\A_i$ is
\emph{$(\alpha,\beta)$-reliable for $\phi$ in $\N$ (for agent $i$)} if
$\alpha,\beta\in[0,1]$ and for all states $s$ and derandomizers $v$,
\begin{itemize}
\item $(\N,s,v)\sat \phi$ implies 
$\mu_s(\YES) \ge \alpha$;
\item $(\N,s,v)\sat\neg \phi$
implies $\mu_s(\YES) \le \beta$.
\end{itemize}
These conditions are equivalent to 
saying $N \sat (\phi \rimp \Pr(X_i\phi)\ge\alpha) \land (\neg \phi \rimp 
\Pr(X_i\phi)\le\beta)$.  
In other words, if
$\phi$ is true at state $s$, then an $(\alpha,\beta)$-reliable
algorithm says $\YES$ to $\phi$ at $s$ with probability at least
$\alpha$
(and hence is right when it answers $\YES$ to query $\phi$ with
probability at least $\alpha$);
on the other hand, if $\phi$ is false, it says $\YES$ with probability
at most $\beta$ 
(and hence is wrong when it answer $\YES$ to query $\phi$ with
probability at most $\beta$).  
The primality testing knowledge algorithm is
$(1,1/2)$-reliable 
for $\pprime$.
The intuition here is that $(\alpha,\beta)$-reliability is a way to
bound the probability that the knowledge algorithm is wrong. The
knowledge algorithm can be wrong in two ways: it can answer 
$\NO$ or $\DONTK$ to a query $\phi$ when $\phi$ is true, and it can
answer $\YES$ to a query $\phi$ when $\phi$ is not true. 
If a knowledge algorithm is $(\alpha,\beta)$-reliable, then the
probability that it answers $\NO$ or $\DONTK$ when the answer should be
$\YES$ is at most $1-\alpha$: the probability that it answers $\YES$
when it should not is at most $\beta$.

\newcommand{\Kpmiphi}{K^{\pm}_i\phi}

We can now capture the relationship between reliable
knowledge algorithms and evidence.
The relationship depends in part on what the agent considers possible.
\pro\label{p:properties}
If $\A_i$ is
$\phi$-complete and 
$(\alpha,\beta)$-reliable for $\phi$ in $\N$
then 
\begin{itemize}
\item[(a)] $\N \sat 
X_i\phi \land \neg K_i \neg \phi\rimp \lEv_i(\phi)\ge\frac{\alpha}{\alpha+\beta}$ 
if $(\alpha,\beta) \ne (0,0)$;
\item[(b)]
$\N \sat X_i\phi\land \neg K_i \neg \phi\rimp \lEv_i(\phi) =  1$ if
$(\alpha,\beta) = (0,0)$; 
\item[(c)] 
$\N \sat \neg X_i\phi\land \neg K_i \phi\rimp 
    \uEv_i(\phi)\le\frac{1-\alpha}{2-(\alpha+\beta)}$
if $(\alpha,\beta) \ne (1,1)$;
\item[(d)] $\N \sat 
\neg X_i\phi\land\neg K_i \phi\rimp 
\uEv_i(\phi) = 0$ if $(\alpha,\beta) = (1,1)$.
\end{itemize}
\epro 

Proposition~\ref{p:properties} becomes interesting in the context of
well-known classes of randomized algorithms \cite{r:motwani95}. 
An $\textbf{RP}$ (random polynomial-time) algorithm is
a polynomial-time randomized algorithm that is
$(1/2,0)$-reliable. 
It thus follows from Proposition~\ref{p:properties} that if $\A_i$ is an
{\bf RP} algorithm, then 
$N \sat (X_i\phi \land \neg K_i \neg \phi \rimp \lEv_i(\phi)=1) \land
(\neg X_i\phi \land \neg K_i \phi \rimp 
\uEv_i(\phi)\leq 1/3)$.
By Proposition~\ref{p:evidence-knowledge}, 
$\lEv_i(\phi)=1\rimp \phi$ is valid, and thus 
we have $N \sat X_i\phi \land \neg K_i \neg \phi \rimp \phi$.
A $\textbf{BPP}$ (bounded-error probabilistic
polynomial-time) algorithm is a
polynomial-time randomized algorithm that is
$(3/4,1/4)$-reliable.
Thus, by Proposition~\ref{p:properties}, if $\A_i$ is a {\bf BPP} algorithm, 
then 
$N \sat (X_i\phi \land \neg K_i \neg \phi \rimp \lEv_i(\phi)\geq 3/4) \land
(\neg X_i\phi \land \neg K_i \phi \rimp 
\uEv_i(\phi)\leq 1/4).$  

Notice that 
Proposition~\ref{p:properties} talks about
the evidence that the knowledge algorithm provides for
$\phi$. Intuitively, we might expect some kind of relationship 
between the evidence for $\phi$ and the evidence for
$\neg\phi$. A plausible relationship would be that high evidence for
$\phi$ implies low evidence for 
$\neg\phi$, and low evidence for $\phi$ implies high evidence for
$\neg\phi$. Unfortunately, given the definitions in this section, this 
is not the case. Evidence for $\phi$ is completely unrelated to
evidence for $\neg\phi$. Roughly speaking, this is because evidence for
$\phi$ is measured by looking at the results of the knowledge
algorithm when queried for $\phi$, and evidence for $\neg\phi$
is measured by looking at the results of the knowledge algorithm when
queried for $\neg\phi$. 
There is nothing in the definition of
a knowledge algorithm that says that the answers of the knowledge
algorithm to queries $\phi$ and $\neg\phi$ need to be related in any
way.

A relationship between evidence for $\phi$ and evidence for $\neg\phi$
can be established by considering knowledge algorithms that are
``well-behaved'' with respect to negation. 
There is a natural way to define the behavior of a knowledge algorithm
on negated formulas. Intuitively, a strategy to evaluate
$\A_i(\neg\phi,\ell,s,v_i)$ is to evaluate $\A_i(\phi,\ell,s,v_i)$, and
returns the negation of the result. There is a choice to be
made in the case when the $\A_i$ returns $\DONTK$ to the query for
$\phi$. One possibility is to return $\DONTK$ to the query for
$\neg\phi$ when the query for $\phi$ returns $\DONTK$; another
possibility is to return $\YES$ is the query for $\phi$ returns
$\DONTK$.  A randomized knowledge algorithm $\A$
\emph{weakly respects negation} if, for all local states $\ell$ and 
derandomizers $v$,
\[ \Ad(\neg\phi,\ell,s,v_i)=\begin{cases}
   \YES & \text{if $\Ad(\phi,\ell,s,v_i)=\NO$}\\
   \NO & \text{if $\Ad(\phi,\ell,s,v_i)=\YES$}\\
   \DONTK & \text{if $\Ad(\phi,\ell,s,v_i)=\DONTK$.}\end{cases}\]
Similarly, a randomized knowledge algorithm $\A$
\emph{strongly respects negation}, if for
all local states $\ell$ and derandomizers $v$,
\[ \Ad(\neg\phi,\ell,s,v_i)=\begin{cases}
   \YES & \text{if $\Ad(\phi,\ell,s,v_i)\ne\YES$}\\
   \NO & \text{if $\Ad(\phi,\ell,s,v_i)=\YES$.}\\\end{cases}\]
Note that if $\A_i$ is $\phi$-complete, then the output of $\A_i$ on
input $\neg\phi$ is the same whether $\A_i$ weakly or strongly
respects negation. 
Say $\A_i$ \emph{respects negation} if it weakly or
strongly respects negation. 
Note that
if $\A_i$ is $\phi$-complete and respects negation, then $\A_i$ is
$\neg\phi$-complete. 

Our first result shows that 
for knowledge algorithms that 
respect negation, reliability for $\phi$
is related to reliability for $\neg\phi$:
\pro\label{p:negreliable}
If $\A_i$ respects negation, is
$\phi$-complete, and 
is
$(\alpha,\beta)$-reliable for $\phi$ in $\N$, then
$\A_i$ is $(\alpha,\beta)$-reliable for $\phi$ in $\N$ if and only if
$\A_i$ is $(1-\beta,1-\alpha)$-reliable for $\neg\phi$ in $\N$. 
\epro

It is easy to check that if 
$\A_i$ 
is $\phi$-complete and respects negation, then 
$X_i\phi\riff\neg X_i\neg\phi$ is a valid formula. 
Combined with
Proposition~\ref{p:properties},
this yields the following results.

\pro\label{p:negproperties}
If $\A_i$ respects negation, is
$\phi$-complete, and 
is
$(\alpha,\beta)$-reliable for $\phi$ in $\N$, then
\begin{itemize}
\item[(a)] $\N \sat X_i\phi\land\neg K_i \neg \phi\rimp
\left(\lEv_i(\phi)\ge\frac{\alpha}{\alpha+\beta}\land 
               \uEv_i(\neg\phi)\le\frac{\beta}{\alpha+\beta}\right)$
if $(\alpha,\beta) \ne (0,0)$;
\item[(b)] $\N \sat X_i\phi\land\neg K_i \neg \phi\rimp 
\left(\lEv_i(\phi)=1\land
               \uEv_i(\neg\phi)=0\right)$ if $(\alpha,\beta) = (0,0)$;
\item[(c)] $\N \sat
X_i\neg\phi\land \neg K_i \phi\rimp
\left(\lEv_i(\neg\phi)\ge\frac{1-\beta}{2-(\alpha+\beta)}\land   
                  \uEv_i(\phi)\le\frac{1-\alpha}{2-(\alpha+\beta)}\right)$
if $(\alpha,\beta) \ne (1,1)$;
\item[(d)] $\N \sat
X_i\neg\phi\land \neg K_i 
\phi\rimp\left(\lEv_i(\neg\phi)\ge\frac{1}{2}\land  
               \uEv_i(\phi)\le\frac{1}{2}\right)$ 
if $(\alpha,\beta) = (1,1)$.
\end{itemize}
\commentout{
If $\A_i$ is $(0,0)$-reliable for $\phi$ in $\N$,
and $\A_i$ strongly respects negation, then the following formulas
are valid in $\N$: 
\begin{gather*}
X_i\phi\rimp \left(\Ev_i(\phi)=1\land
               \Ev_i(\neg\phi)=0\right),\\
X_i\neg\phi\rimp\left(\Ev_i(\neg\phi)\ge\frac{1}{2}\land
                  \Ev_i(\phi)\le\frac{1}{2}\right).
\end{gather*}
If $\A_i$ is $(1,1)$-reliable for $\phi$ in $\N$
and $\A_i$ strongly respects negation, then the following formulas
are valid in $\N$: 
\begin{gather*}
X_i\phi\rimp \left(\Ev_i(\phi)\ge\frac{1}{2}\land
               \Ev_i(\neg\phi)\le\frac{1}{2}\right),\\
X_i\neg\phi\rimp\left(\Ev_i(\neg\phi)=1\land
                  \Ev_i(\phi)=0\right).
\end{gather*}
}
\epro

\section{Conclusion}\label{s:conc}

\commentout{
The first goal of this paper is to define the semantics of
algorithmic knowledge in the presence of randomized knowledge
algorithms. This is done by essentially derandomizing the knowledge
algorithms, supplying them with an extra argument representing the
random information (for instance, outcomes of coin tosses if the
algorithm makes random choices by tossing coins). 
Using Proposition~\ref{p:properties}, 
we can 
precisely characterize the contribution of
algorithmic knowledge using randomized knowledge 
algorithms in terms of evidence.
Note that this is the evidence provided by a single query to the knowledge
algorithm. 
Of course, it is always possible to increase the reliability of a
knowledge algorithm, by essentially iterating it.
(Alternatively, we can 
compute the cumulative evidence for
$\phi$ provided by $k$ queries to the knowledge algorithm, either
directly by constructing an appropriate evidence space (the observations
are now sequences of $\YES$, $\NO$, and $\DONTK$ of length $k$) or by
using {\em Dempster's Rule of Combination} \cite{r:shafer76}.
See \cite{r:halpern03b} for a discussion of combining the evidence of a
number of observations, as well combining evidence with prior
probabilities.)  
}
The goal of this paper is to understand what the evidence provided by a
knowledge algorithm tells us.  
To take an example from
security, consider an enforcement mechanism used to detect and react
to intrusions in a system. Such an enforcement mechanism uses
algorithms that analyze the behavior of users and attempt to recognize
intruders. While the algorithms may sometimes be wrong, they are typically
reliable, in our sense, with some associated probabilities. 
Clearly the mechanism wants to make sensible decisions based on this
information.   How should it do this?
What actions should the system take based on a report that a user is an
intruder?  

If we have a probability on the hypotheses, evidence 
can be used to update this probability. 
More precisely,
as shown in \cite{HF2}, evidence can be
viewed as a function from priors to posteriors.  For example, if the
(cumulative) evidence for $n$ being a prime is $\alpha$ and the prior
probability that $n$ is prime is $\beta$, then 
a straightforward application of 
Bayes' rule tells us that the posterior probability of $n$ being prime
(that is, the probability of $n$ being prime in light of the evidence) 
is $(\alpha \beta)/ (\alpha\beta + (1-\alpha)(1-\beta))$.%
\footnote{The logic in this paper only considers the evidence provided by 
  a single knowledge algorithm at a single point in time. 
  In general, evidence from multiple sources can be accumulated over
  time and combined.  
  Our companion paper \cite{r:halpern03b} discusses a more general logic
  in which the combination of evidence can be expressed.}
Therefore,
if we have a prior probability on the hypotheses, including the
formula $\phi$, then we can decide to perform an action when the
posterior probability of $\phi$ is high enough. 
(A similar interpretation holds for the evidence expressed
by $\wlow_{\cE}$
and $\wup_{\cE}$; 
we hope to report on this topic in future work.)
However, what can we
do when there is no probability distribution on the hypotheses, as in
the primality example at the beginning of this section? The 
probabilistic interpretation of evidence still gives us a guide for
decisions. As before, we assume that if the posterior probability of
$\phi$ is high enough, we will act as if $\phi$ holds. The problem, of
course, is that we do not have a prior probability. However, the
evidence tells us what prior probabilities we must be willing to
assume for the posterior probability to be high enough.
For example, a $\YES$ from a $(.999,.001)$-reliable
algorithm for $\phi$ says that as long as the prior probability of
$\phi$ is at least $.01$, then the posterior is at least .9.  This may
be sufficient assurance for an agent to act.

Of course, it is also possible to treat evidence as primitive, and
simply decide to act is if the hypothesis for which there is more
evidence, or for the hypothesis for which evidence is above a certain
threshold is true. It would in fact be of independent interest to study the
properties of a theory of decisions based on a primitive notion of
evidence.  We leave this to future work.

\section*{Acknowledgments}

This work was mainly performed while the second author was at Cornell
University. 
A preliminary version of this paper appeared in the \emph{Proceedings
of the Ninth Conference on Theoretical Aspects of Rationality and
Knowledge}, pp. 118--130, 2003.
Thanks to Tugkan Batu and Hubie Chen for discussions.
Authors supported in part by NSF under
grant CTC-0208535, by ONR under grants N00014-00-1-03-41 and
N00014-01-10-511, by the DoD Multidisciplinary University Research
Initiative (MURI) program administered by the ONR under grant
N00014-01-1-0795, and by AFOSR under grant F49620-02-1-0101.

\appendix

\section{Proofs}

\begin{oldtheorem}{p:deterministic}
\pro
Let $\N=(S,\pi,L_1,\dots,L_n,\Ad_1,\dots,\Ad_n,\nu)$ be a
probabilistic algorithmic knowledge, with $\A_1,\dots,\A_n$
deterministic. Let $M=(S,\pi,L_1,\dots,L_n,\A_1,\dots,\A_n)$. If there
are no occurrences of $\Pr$ in $\phi$, then for all $s\in S$ and all
$v\in V$, $(\N,s,v)\sat\phi$ if and only if $(M,s)\sat\phi$. 
\epro
\end{oldtheorem}
\prf
The key observation here is that if a knowledge algorithm $\A$ is
deterministic, then for all $v\in V$,
$\A^d(\phi,\ell,s,v_i)=\A(\phi,\ell,s)$. The result then follows
easily by induction on the structure of $\phi$. If $\phi$ is $p$, then
$(\N,s,v)\sat p$ if and only if $\pi(s)(p)=\true$ if and only if
$(M,s)\sat p$. If $\phi$ is $\psi_1\land\psi_2$, then
$(\N,s,v)\sat\psi_1\land\psi_2$ if and only if $(\N,s,v)\sat\psi_1$
and $(\N,s,v)\sat\psi_2$ if and only if $(M,s)\sat\psi_1$ and
$(M,s)\sat\psi_2$ (by the induction hypothesis) if and only if
$(M,s)\sat\psi_1\land\psi_2$. If $\phi$ is $\neg\psi$, then
$(\N,s,v)\sat\neg\psi$ if and only if $(\N,s,v)\not\sat\psi$ if and
only $(M,s)\not\sat\psi$ (by the induction hypothesis) if and only if
$(M,s)\sat\neg\psi$. If $\phi$ is $K_i\psi$, suppose that
$(\N,s,v)\sat K_i\psi$, that is, for all $v'\in V$ and all $t\sim_i
s$, $(\N,t,v')\sat \psi$; by the induction hypothesis, this means that
for all $t\sim_i s$, $(M,t)\sat\psi$, that is, $(M,s)\sat K_i \psi$.
Conversely, suppose that $(M,s)\sat K_i\psi$, so that for all
$t\sim_i s$, $(M,t)\sat\psi$; by the induction hypothesis, for every
$t\sim_i s$, we have $(\N,t,v')\sat \psi$ for all $v'\in V$, and thus
$(\N,s,v)\sat K_i\psi$. If $\phi$ is $X_i\psi$, then $(\N,s,v)\sat
X_i\psi$ if and only if $\A_i^d(\psi,L_i(s),s,v_i)=\YES$ if and
only if $\A_i(\psi,L_i(s),s)=\YES$ (since $\A_i$ is deterministic) if
and only if $(M,s)\sat X_i\psi$.
\eprf

\begin{oldtheorem}{p:xfree}
\pro
Let $\N=(S,\pi,L_1,\dots,L_n,\Ad_1,\dots,\Ad_n,\nu)$ be a
probabilistic algorithmic knowledge structure, and let
$M=(S,\pi,L_1,\dots,L_n,\A'_1,\dots,\A'_n)$ be an algorithmic
knowledge structure where $\A'_1,\dots,\A'_n$ are arbitrary
deterministic knowledge algorithms. If there are no occurrences of
$X_i$ and $\Pr$ in $\phi$, then for all $s\in S$ and all $v\in V$,
$(\N,s,v)\sat\phi$ if and only if $(M,s)\sat\phi$.
\epro
\end{oldtheorem}
\prf
This result in fact follows from the proof of
Proposition~\ref{p:deterministic}, since the only use of the 
assumption that knowledge algorithms are deterministic is in the
inductive step for subformulas of the form $X_i\psi$. 
\eprf

\begin{oldtheorem}{p:probalgk}
\pro
Suppose that $\N = (S,\pi,L_1,\ldots,L_n,\A_1^d,\ldots,\A_n^d,\nu)$ is
a probabilistic algorithmic knowledge security structure with an adversary as
agent $i$ and that $\A_i = \ADYgi$. Let $K$ be the number of distinct
keys used in the messages in the adversary's local state $\ell$ (i.e.,
the number of keys used in the messages that the adversary has 
intercepted at a state $s$ with $L_i(s)=\ell$).  Suppose that $K/\abs{\cK} <
1/2$ and that $\nu$ is the uniform distribution on sequences of coin
tosses.  If $(\N,s,v) \sat \neg K_i X_i (\sees_i(\msg))$, then
$(\N,s,v) \sat \Pr(X_i (\sees_i(\msg))) < 1 - e^{-2\ng K/\abs{\cK}}$.
Moreover, 
if $(\N,s,v) \sat X_i(\sees_i(\msg))$ then $(\N,s,v) \sat
\sees_i(\msg)$.
\epro
\end{oldtheorem}
\prf It is not hard to show that the $\ng$ keys that the adversary
guesses do no good at all if none of them match a key used in a
message intercepted by the adversary.  By assumption, $K$ keys are
used in messages intercepted by the adversary.  The probability that a
key chosen at random is one of these $K$ is $K/\abs{\cK}$, since there are $\abs{\cK}$
keys altogether.  Thus, the probability that a key chosen at random is
not one of these $K$ is $1-(K/\abs{\cK})$.  The probability that none of the
$\ng$ keys chosen at random is one of these $K$ is therefore $(1 -
(K/\abs{\cK}))^\ng$.  We now use some standard approximations.  Note that $(1 -
(K/\abs{\cK}))^\ng = e^{\ng\ln(1 - (K/\abs{\cK})}$, and
$$\ln(1-x) = -x - x^2/2 - x^3/3 - \cdots > - x - x^2 - x^3 - \cdots =
-x/(1-x).$$
Thus, if $0 < x < 1/2$, then 
$\ln(1-x) > -2x$.  It follows that if $K/\abs{\cK} < 1/2$, then $e^{\ng\ln(1 -
(K/\abs{\cK}))} > e^{-2\ng K/\abs{\cK}}$. 
Since the probability that a key chosen at random does not help to
compute algorithmic knowledge is
greater than $e^{-2\ng K/\abs{\cK}}$, the probability that it helps is less than
$1 - e^{-2\ng K/\abs{\cK}}$.

Soundness of $\A_i$ with respect to $\sees_i(\msg)$ follows from 
Proposition~\ref{p:dolevyao} (since soundness follows for arbitrary
$\mathit{initkeys}(\ell)\subseteq\cK$). 
\eprf

\begin{oldtheorem}{p:defined}
\pro
For all probabilistic algorithmic knowledge structures $N$,
agents $i$, formulas $\phi$, and local states $\ell$ of agent $i$ that
arise in $N$, if $\ob$ is a possible output of $i$'s knowledge algorithm 
$\Ad_i$ in local state $\ell$ on input $\phi$, then there exists a
probability measure $\mu \in \cF_{\ell,\phi}(\phi) \union 
\cF_{\ell,\phi}(\neg \phi)$ such that $\mu(\ob) > 0$.
\epro
\end{oldtheorem}
\prf
Suppose that $\ell$ is a local state of agent $i$ that arises in $N$ and
$\ob$ is a possible output of $\Ad_i$ in local state $\ell$ on input
$\phi$. Thus, there exists a state $s$ and derandomizer $v$ such that
$\Ad_i(\phi,L_i(s),s,v_i)=\ob$.  By assumption, $\mu_s(\ob) > 0$.
\eprf

\begin{oldtheorem}{p:evidence-knowledge}
\pro
For all probabilistic algorithmic knowledge structures $\N$, 
we have $$N \sat \lEv_i(\phi)=1\rimp \phi.$$
\epro
\end{oldtheorem}
\prf
If $(\N,s,v)\sat\lEv_i(\phi)=1$, then
$\wlow_{\cE_{\A_i,\phi,L_i(s)}}(\Ad_i(\phi,L_i(s),v_i),\phi)=1$.
By definition of $\wlow_{\cE_{\A_i,\phi,L_i(s)}}$, this implies that
for all $s'\in S_{L_i(s),\neg\phi}$, 
$\mu_{s'}(\Ad_i(\phi,L_i(s),s,v_i))=0$. By our assumption about
derandomizers, this means that there is no state $s'$ and derandomizer
$v'$ such that 
$L_i(s')=L_i(s)$ and $(\N,s',v')\sat\neg\phi$ where
$\Ad_i(\phi,L_i(s),s',v'_i)=\Ad_i(\phi,L_i(s),s,v_i)$. Thus, we cannot 
have $(\N,s,v)\sat\neg\phi$. 
Hence, $(\N,s,v)\sat\phi$, as required. 
\eprf

\commentout{
To prove Propositions~\ref{p:properties} and \ref{p:negproperties},
the following lemma, alluded to in the text, is useful. It describes
the relationship between the reliability of a knowledge algorithm and
the properties of probability measures in $\cP_\phi$ and
$\cP_{\neg\phi}$ that appear  
in the evidence space $\cE_{\A_i,\phi,\ell}$, as
defined in Section~\ref{s:reliable}. 

\begin{lem}\label{l:evidence}
If $\A_i$ is $\phi$-complete and $(\alpha,\beta)$-reliable for $\phi$
in $\N$, then for all states $s,s'$ with $L_i(s)=L_i(s')$,
$(\N,s)\sat\phi$, and $(\N,s')\sat\neg\phi$, we have
$\mu_s(\YES)\ge\alpha$, $\mu_s(\NO)\le 1-\alpha$,
$\mu_{s'}(\YES)\le\beta$, and $\mu_{s'}(\NO)\ge 1-\beta$.
\end{lem}
\prf
This is immediate from the definitions. By assumption, $s\in
S_{L_i(s),\phi}$ and $s'\in S_{L_i(s),\neg\phi}$. By the definition of 
reliability, since for every $v$, $(\N,s,v)\sat\phi$, then
$\mu_s(\YES)=\nu(\{v'\mid \Ad_i(\phi,L_i(s),s,v'_i)=\YES\})\ge
\alpha$, and thus by $\phi$-completeness, $\mu_s(\NO)=1-\mu_s(\YES)\le
1-\alpha$. Similarly, since for every $v$, $(\N,s',v)\sat\neg\phi$, then
$\mu_{s'}(\YES)=\nu(\{v'\mid \Ad_i(\phi,L_i(s),s',v'_i)=\YES\})\le
\beta$, and thus by $\phi$-completeness,
$\mu_{s'}(\NO)=1-\mu_{s'}(\YES)\ge 1-\beta$. 
\eprf
}
The following lemma gives an algebraic relationship that is 
useful in the proofs of Propositions~\ref{p:properties}
and \ref{p:negproperties}. 
\begin{lem}\label{l:ineq}
Suppose that $x,y,a,b$ are real numbers in $[0,1]$ such that $x$ and $y$
are not both $0$, and $a$ and $b$ are not both $0$. If $x\ge a$ and
$y\le b$, then $x/(x+y)\ge a/(a+b)$ and $y/(x+y)\le b/(a+b)$. 
\end{lem}
\prf
Note that $x(a+b)=xa+xb\ge xa+ay = a(x+y)$, so that $x/(x+y)\ge
a/(a+b)$. Similarly, $y(a+b)=ya+yb\le xb+yb=b(x+y)$, so that
$y/(x+y)\le b(a+b)$.
\eprf

\begin{oldtheorem}{p:properties}
\pro
If $\A_i$ is $\phi$-complete and $(\alpha,\beta)$-reliable for $\phi$ in
$\N$ then 
\begin{itemize}
\item[(a)] $\N \sat 
X_i\phi \land \neg K_i \neg \phi\rimp \lEv_i(\phi)\ge\frac{\alpha}{\alpha+\beta}$ 
if $(\alpha,\beta) \ne (0,0)$;
\item[(b)]
$\N \sat X_i\phi\land \neg K_i \neg \phi\rimp \lEv_i(\phi) =  1$ if
$(\alpha,\beta) = (0,0)$; 
\item[(c)] $\N \sat \neg X_i\phi\land \neg K_i \phi\rimp 
    \uEv_i(\phi)\le\frac{1-\alpha}{2-(\alpha+\beta)}$
if $(\alpha,\beta) \ne (1,1)$;
\item[(d)] $\N \sat 
\neg X_i\phi\land\neg K_i \phi\rimp 
\uEv_i(\phi) = 0$ if $(\alpha,\beta) = (1,1)$.
\end{itemize}
\epro 
\end{oldtheorem}
\prf
For part (a), suppose that $(\alpha,\beta) \ne (0,0)$
and that $(\N,s,v)\sat X_i\phi\land\neg K_i \neg \phi$.
Then $\A^d_i(\phi,L_i(s),s,v_i)=\YES$ and there exists some $s' \sim_i
s$ such that $(\N,s',v) \sat \phi$.  By the latter fact,
$S_{L_i(s),\phi} \ne \emptyset$.  If $S_{L_i(s),\neg \phi} = \emptyset$,
then it is easy to see that $(\N,s,v) \sat \lEv_i(\phi) =  1$.  
If $S_{L_i(s),\neg \phi} \ne \emptyset$,
let $s',s''$ be two arbitrary states in $S_{L_i(s),\phi}$ and
$S_{L_i(s),\neg\phi}$, respectively. 
Since $\Ad_i$ is $(\alpha,\beta)$ reliable, 
$\mu_{s',\phi}(\YES)\ge\alpha$ and
$\mu_{s'',\phi}(\YES)\le\beta$. 
Therefore, by Lemma~\ref{l:ineq}, we have 
$$\frac{\mu_{s',\phi}(\YES)}{\mu_{s',\phi}(\YES)+\mu_{s'',\phi}(\YES)}
\ge\frac{\alpha}{(\alpha+\beta)}.$$ 
Since $s'$ and $s''$ were arbitrary, 
it follows that
$\wlow_{\cE_{\A_i,\phi,L_i(s)}}(\YES,\phi)\ge
\alpha/(\alpha+\beta)$. 
Thus, we have $(\N,s,v)\sat\lEv_i(\phi)\ge\alpha/(\alpha+\beta)$. 
Since $s$ and $v$ were arbitrary, we have 
$$\N\sat
X_i\phi\land\neg K_i\neg\phi\rimp\lEv_i(\phi)\ge\frac{\alpha}{\alpha+\beta}.$$

For part (b), suppose that $(\alpha,\beta)=(0,0)$, and that
$(\N,s,v)\sat X_i\phi\land\neg K_i \neg \phi$.  Then
$\A^d_i(\phi,L_i(s),s,v_i)=\YES$ and there exists some $s' \sim_i s$
such that $(\N,s',v) \sat \phi$.  By the latter fact, $S_{L_i(s),\phi}
\ne \emptyset$.  If $S_{L_i(s),\neg \phi} = \emptyset$, then it is
easy to see that $(\N,s,v) \sat \lEv_i(\phi) = 1$.  If $S_{L_i(s),\neg
\phi} \ne \emptyset$, consider all pairs of states $s',s''$ with
$s'\in S_{L_i(s),\phi}$ and $s''\in S_{L_i(s),\neg\phi}$.  Since
$\Ad_i$ is $(0,0)$ reliable, $\mu_{s',\phi}(\YES)\ge 0$ and
$\mu_{s'',\phi}(\YES)= 0$. For
such a pair $s',s''$, either 
$\mu_{s',\phi}(\YES)=0$, in which case
$\mu_{s',\phi}(\YES)+\mu_{s'',\phi}(\YES)=0$, and the pair $s',s''$
does not contribute to $\cW_{\cE_{\A_i,\phi,L_i(s)}}(\YES,\phi)$; or
$\mu_{s',\phi}(\YES)=\alpha >0$ (and by Proposition~\ref{p:defined},
at least one such state $s'$ always exists), 
in which an argument similar to
part (a) applies and, by Lemma~\ref{l:ineq}, we have
\[\frac{\mu_{s',\phi}(\YES)}{\mu_{s',\phi}(\YES)+\mu_{s'',\phi}(\YES)}
\ge\frac{\alpha}{\alpha} = 1.\]
Thus, $\cW_{\cE_{\A_i,\phi,L_i(s)}}(\YES,\phi)=\{1\}$, 
$\wlow_{\cE_{\A_i,\phi,L_i(s)}}(\YES,\phi)=1$, and
$(\N,s,v)\sat\lEv_i(\phi)=1$.  Since $s$ and $v$ were arbitrary, we
have $\N\sat X_i\phi\land\neg K_i \neg\phi\rimp\lEv_i(\phi)=1$.

For part (c), suppose that $(\alpha,\beta) \ne (1,1)$ 
and that
$(\N,s,v)\sat \neg X_i\phi\land \neg K_i \phi$.
Thus,
$\A^d_i(\phi,L_i(s),v_i)=\NO$ (since $\A_i$ is $\phi$-complete)
and there exists some $s' \sim_i s$ such that $(\N,s',v) \sat \neg \phi$.
By the latter fact,
$S_{L_i(s),\neg \phi} \ne \emptyset$.  If $S_{L_i(s), \phi} = \emptyset$,
then it is easy to see that $(\N,s,v) \sat \uEv_i(\phi) =  0$.  
If $S_{L_i(s), \phi} \ne \emptyset$,
let $s',s''$ be two arbitrary states in $S_{L_i(s),\phi}$ and
$S_{L_i(s),\neg\phi}$, respectively. 
Since $\A^d_i$ is $\phi$-complete and $(\alpha,\beta)$-reliable, 
we have $\mu_{s',\phi}(\NO)\le
1-\alpha$ and 
$\mu_{s'',\phi}(\NO)\ge 1-\beta$. 
(This is where we use $\phi$-completeness; otherwise, the best we can
say is that $\mu_{s'',\phi}(\NO)+\mu_{s'',\phi}(\DONTK)\ge 1-\beta$.)
Therefore, by Lemma~\ref{l:ineq}, we have 
$$\frac{\mu_{s',\phi}(\NO)}{\mu_{s',\phi}(\NO)+\mu_{s'',\phi}(\NO)} \le
\frac{1-\alpha}{2-(\alpha+\beta)}.$$ 
Since $s'$ and $s''$ were arbitrary, 
we have $(\N,s,v)\sat\uEv_i(\phi)\le 1-\alpha/(2-(\alpha+\beta))$.
Since $s$ and $v$ were arbitrary, we have 
$$\N\sat
X_i\phi\land\neg K_i \phi\rimp\uEv_i(\phi)\le
\frac{1-\alpha}{2-(\alpha+\beta)}.$$ 

The proof of part (d) is similar to that of (b), and is left to the
reader. 
\eprf

\begin{oldtheorem}{p:negreliable}
\pro
If $\A_i$ respects negation, is $\phi$-complete, and is
$(\alpha,\beta)$-reliable for $\phi$ in $\N$, then $\A_i$ is
$(\alpha,\beta)$-reliable for $\phi$ in $\N$ if and only if $\A_i$ is
$(1-\beta,1-\alpha)$-reliable for $\neg\phi$ in $\N$.
\epro
\end{oldtheorem}
\prf
This is almost immediate from the definitions.
Suppose that $\A_i$ respects negation, is $\phi$-complete, and is
$(\alpha,\beta)$-reliable for $\phi$ in $\N$.  
Consider the reliability of $\A_i$ with respect to $\neg\phi$. If
$(\N,s,v)\sat\neg\phi$, then
$$\mu_{s,\neg \phi}(\YES) = 1 - \mu_{s,\neg \phi}(\NO)  = 1 -
\mu_{s,\phi}(\YES) \ge 1-\beta.$$
Similarly, if
$(\N,s,v)\sat\phi$, then 
$\mu_{s,\neg \phi}(\YES) \le 1 - \alpha$.
Thus, $\A_i$ is
$(1-\beta,1-\alpha)$-reliable for $\neg\phi$ in $\N$. 

\commentout{
Conversely, suppose that $\A_i$ is $(1-\beta,1-\alpha)$-reliable for
$\neg\phi$ in $\N$, and consider the reliability of $\A_i$ with
respect to $\phi$. If $(\N,s,v)\sat\phi$, then 
\begin{align*}
& \nu(\{v' \mid\A^d_i(\phi,L_i(s),s,v'_i)=\YES\})\\
& \quad = 1-\nu(\{v' \mid\A^d_i(\phi,L_i(s),s,v'_i)\ne\YES\})\\
& \quad = 1-\nu(\{v' \mid\A^d_i(\phi,L_i(s),s,v'_i)=\NO\})\quad\text{by $\phi$-completeness}\\
& \quad = 1-\nu(\{v' \mid\A^d_i(\neg\phi,L_i(s),s,v'_i)=\YES\})\\
& \quad \ge 1-(1-\alpha)\\
& \quad = \alpha.
\end{align*}
Similarly, if $(\N,s,v)\sat\neg\phi$, then
\begin{align*}
&\nu(\{v' \mid \A^d_i(\phi,L_i(s),s,v'_i)=\YES\})\\
& \quad = 1-\nu(\{v' \mid\A^d_i(\phi,L_i(s),s,v'_i)\ne\YES\})\\
& \quad = 1-\nu(\{v' \mid\A^d_i(\phi,L_i(s),s,v'_i)=\NO\})\quad\text{by $\phi$-completeness}\\
& \quad = 1-\nu(\{v' \mid\A^d_i(\neg\phi,L_i(s),s,v'_i)=\YES\})\\
& \quad \le 1-(1-\beta)\\
& \quad =\beta.
\end{align*}
Thus, 
$\A_i$ is $(\alpha,\beta)$-reliable for $\phi$ in $\N$. 
}
An identical argument (replacing $\phi$ by $\neg \phi$, and
$(\alpha,\beta)$ by $(1-\alpha, 1-\beta)$), shows that if $\A_i$ is
$(1-\beta,1-\alpha)$-reliable for $\neg\phi$ in $\N$ then 
$\A_i$ is $(\alpha,\beta)$-reliable for $\phi$ in $\N$.  We leave
details to the reader.
\eprf

To prove Proposition \ref{p:negproperties}, we need a preliminary
lemma, alluded to in the text.

\begin{lem}\label{l:equivneg}
If $\N$ is a probabilistic algorithmic knowledge structure where agent
$i$ uses a knowledge algorithm $\A_i$ that is $\phi$-complete and that
respects negation, then 
$N \sat X_i\phi\riff \neg X_i\neg\phi$.
\end{lem}
\prf
Let $s$ be a state of $\N$ and let $v$ be a derandomizer.  If
$(\N,s,v)\sat X_i\phi$, then $\Ad_i(\phi,L_i(s),s,v_i)=\YES$. Since
$\A_i$ respects negation and is $\phi$-complete, this implies that
$\Ad_i(\neg\phi,L_i(s),s,v_i)=\NO$ 
($\Ad_i$ cannot return $\DONTK$ since it is $\phi$-complete) and hence
that $(\N,s,v)\not\sat 
X_i\neg\phi$, so $(\N,s,v)\sat\neg X_i\neg\phi$. Thus, $(\N,s,v)\sat
X_i\phi\rimp\neg X_i\neg\phi$. Since $s$ and $v$ were arbitrary, we
have that $\N\sat X_i\phi\rimp\neg X_i\neg\phi$. Conversely, let $s$
be a state of $\N$ and let $v$ be a derandomizer. If $(\N,s,v)\sat\neg
X_i\neg\phi$, then $(\N,s,v)\not\sat X_i\neg\phi$, that is,
$\Ad_i(\neg\phi,L_i(s),s,v_i)\ne\YES$. Since $A_i$ is $\phi$-complete
and respects negation, $A_i$ is $\neg\phi$-complete, so it must be the
case that $\Ad_i(\neg\phi,L_i(s),s,v_i)=\NO$. Therefore,\break
$\Ad_i(\phi,L_i(s),s,v_i)=\YES$, and $(\N,s,v)\sat X_i\phi$. Since $s$
and $v$ were arbitrary, we have that $\N\sat X_i\phi$.
\eprf

\begin{oldtheorem}{p:negproperties}
\pro
If $\A_i$ respects negation, is $\phi$-complete,
and is $(\alpha,\beta)$-reliable for $\phi$ in $\N$, then
\begin{itemize}
\item[(a)] $\N \sat X_i\phi\land\neg K_i\neg\phi\rimp
\left(\lEv_i(\phi)\ge\frac{\alpha}{\alpha+\beta}\land 
               \uEv_i(\neg\phi)\le\frac{\beta}{\alpha+\beta}\right)$
if $(\alpha,\beta) \ne (0,0)$;
\item[(b)] $\N \sat X_i\phi\land\neg K_i \neg\phi \rimp \left(\lEv_i(\phi)=1\land
               \uEv_i(\neg\phi)=0\right)$ if $(\alpha,\beta) = (0,0)$;
\item[(c)] $\N \sat
X_i\neg\phi\land \neg K_i \phi \rimp\left(\lEv_i(\neg\phi)\ge\frac{1-\beta}{2-(\alpha+\beta)}\land 
                  \uEv_i(\phi)\le\frac{1-\alpha}{2-(\alpha+\beta)}\right)$
if $(\alpha,\beta) \ne (1,1)$;
\item[(d)] $\N \sat X_i\neg\phi\land\neg K_i \phi\rimp\left(\lEv_i(\neg\phi)\ge\frac{1}{2}\land
                  \uEv_i(\phi)\le\frac{1}{2}\right)$ 
if $(\alpha,\beta) = (1,1)$.
\end{itemize}
\epro
\end{oldtheorem}
\prf
Suppose that $\A_i$ is $(\alpha,\beta)$-reliable for $\phi$ in $\N$.
Since $\A_i$ is $\phi$-complete and respects negation, by
Proposition~\ref{p:negreliable}, $\A_i$ is
$(1-\beta,1-\alpha)$-reliable for $\neg\phi$ in $\N$.  For part (a),
suppose that $(\alpha,\beta)\ne (0,0)$.  Let $s$ be a state of $\N$
and let $v$ be a derandomizer. By Proposition~\ref{p:properties}
applied to $\phi$, 
$$(\N,s,v)\sat
X_i\phi\land\neg K_i\neg\phi\rimp\Ev_i(\phi)\ge \frac{\alpha}{\alpha+\beta}.$$ By
Lemma~\ref{l:equivneg}, $(\N,s,v)\sat X_i\phi\rimp \neg
X_i\neg\phi$. By Proposition~\ref{p:properties} applied to $\neg\phi$,
$(\N,s,v)\sat\neg
X_i\neg\phi\land\neg K_i\neg\phi\rimp\Ev_i(\neg\phi)\le(1-(1-\beta))/(1-(1-\alpha)+1-(1-\beta)),$
that is, $(\N,s,v)\sat\neg
X_i\neg\phi\land\neg K_i\neg\phi\rimp\Ev_i(\neg\phi)\le
\beta/(\alpha+\beta)$. Putting this together, we get 
\[(\N,s,v)\sat
X_i\phi\land\neg
K_i\neg\phi\rimp\left(\Ev_i(\phi)\ge\frac{\alpha}{\alpha+\beta}\land
\Ev_i(\neg\phi)\le\frac{\beta}{\alpha+\beta}\right).\]    
Since $s$ and $v$ are arbitrary, \[\N\sat
X_i\phi\land\neg
K_i\neg\phi\rimp\left(\Ev_i(\phi)\ge \frac{\alpha}{\alpha+\beta}
\land\Ev_i(\neg\phi)\le \frac{\beta}{\alpha+\beta}\right).\]

For part (c), suppose that $(\alpha,\beta) \ne (1,1)$.  By
Proposition~\ref{p:properties} applied to $\neg\phi$, 
$(\N,s,v)\sat X_i\neg\phi\land\neg
K_i\phi\rimp\Ev_i(\phi)\ge(1-\beta)/(2-(\alpha+\beta))$. By 
Lemma~\ref{l:equivneg}, 
$(\N,s,v)\sat X_i\neg\phi\rimp \neg
X_i\phi$. By Proposition~\ref{p:properties} applied to
$\phi$, 
$$(\N,s,v)\sat\neg
X_i\phi\land\neg K_i \phi\rimp\Ev_i(\phi)\le
\frac{1-\alpha}{2-(\alpha+\beta)}.$$ 
Putting this together, we get \[(\N,s,v)\sat
X_i\neg\phi\land\neg
K_i\phi\rimp\left(\Ev_i(\neg\phi)\ge\frac{1-\beta}{2-(\alpha+\beta)}\land
\Ev_i(\phi)\le \frac{1-\alpha}{2-(\alpha+\beta)}\right).\] 
Since $s$ and $v$ are arbitrary, \[\N\sat
X_i\neg\phi\land\neg
K_i\phi\rimp\left(\Ev_i(\neg\phi)\ge \frac{1-\beta}{2-(\alpha+\beta)}
\land\Ev_i(\phi) \le \frac{1-\alpha}{2-(\alpha+\beta)}\right).\] 

We leave the proof of parts (b) and (d) to the reader.
\eprf

\bibliographystyle{chicagor}
\bibliography{riccardo2,z,joe}

\end{document}